%% file: 285.tex
\documentclass[runningheads]{llncs}
\usepackage{graphicx}

\usepackage{tikz}
\usepackage{comment}
\usepackage{amsmath,amssymb} 
\usepackage{color}
\usepackage{comment}
\usepackage{color}
\usepackage{siunitx}
\usepackage{booktabs}
\usepackage{bm}
\usepackage{mathtools}
\usepackage{xcolor}
\usepackage{floatrow}
\usepackage{multirow}
\usepackage{wrapfig,lipsum,booktabs}
\usepackage{caption}
\usepackage{subcaption}
 \usepackage{sidecap}
\usepackage{xspace}
\usepackage{textcomp}
\usepackage{relsize}
\usepackage{amssymb}
\usepackage{pifont}

\usepackage[accsupp]{axessibility}  

\definecolor{citecolor}{RGB}{59,155,85}
\usepackage[pagebackref,breaklinks,colorlinks]{hyperref}

\usepackage{tikz}  
\usepackage{appendix}
\usepackage{mathtools, nccmath} 

\input{macros}

\begin{document}
\mainmatter

\title{\abbrev: An Improved Tracker For
Panoptic LIDAR Segmentation Data}

\titlerunning{\abbrev}
%

\author{Nirit Alkalay \and Roy Orfaig \and Ben-Zion Bobrovsky}

\authorrunning{N. Alkalay et al.}
%
\institute{School of Electrical Engineering, Tel-Aviv University
\\
\email{{nirit.alkalay@gmail.com},  
\{royorfaig,bobrov\}@tauex.tau.ac.il
}}

\maketitle

\begin{abstract}
4D panoptic LiDAR segmentation is essential for scene understanding in autonomous driving and robotics, combining semantic and instance segmentation with temporal consistency. Current methods, like 4D-PLS and 4D-STOP, use a
tracking-by-detection methodology, employing deep learning networks to perform semantic and instance segmentation on each frame. To maintain temporal consistency, large-size instances detected in the current frame are compared and associated with instances within a temporal window that includes the current and preceding frames. However, their reliance on short-term instance detection, lack of motion estimation, and exclusion of small-sized instances lead to frequent identity switches and reduced tracking performance.  We address these issues with the \textbf{NextStop}\footnote{The source code is available at: \url{https://github.com/AIROTAU/NextStop}} tracker, which integrates Kalman filter-based motion estimation, data association, and lifespan management, along with a tracklet state concept to improve prioritization. Evaluated using the LiDAR Segmentation and Tracking Quality (LSTQ) metric on the SemanticKITTI validation set, NextStop demonstrated enhanced tracking performance, particularly for small-sized objects like \textit{people} and \textit{bicyclists}, with fewer ID switches, earlier tracking initiation, and improved reliability in complex environments.
\end{abstract}


\input{sections/intro}

\input{sections/related_work}

\input{sections/method}

\input{sections/experiments}

\input{sections/conclusion}

\clearpage
%
%
\bibliographystyle{splncs04}
\bibliography{abbrev_short, 285}

\clearpage


\begin{center}
\Large\textbf{Supplementary Material} \\
\end{center}


\appendix
\input{sections/supplementary}

\end{document}

%% file: macros.tex
\usepackage{xspace}
\usepackage[export]{adjustbox}
\usepackage[abs]{overpic}
\usetikzlibrary{arrows.meta}

\usepackage{bold-extra}
\usepackage{cite}

\makeatletter
\newcommand*{\etc}{%
    \@ifnextchar{.}%
        {etc}%
        {etc.\@\xspace}%
}

\definecolor{notetext}{rgb}{0.7,0,0}

\definecolor{ubpubColor}{rgb}{0.43, 0.5, 0.5}
\definecolor{backboneColor}{rgb}{0.423, 0.325, 0.365}
\definecolor{fpnColor}{rgb}{0.255, 0.498, 0.416}

\newcommand{\PAR}[1]{\vskip4pt \noindent {\bf #1~}}

\newcolumntype{P}[1]{>{\centering\arraybackslash}p{#1}}

\newcommand{\abbrev}{NextStop}

\newcommand{\Hquad}{\hspace{0.5em}} 



%% file: sections/intro.tex
\section{Introduction}


In the field of computer vision, scene understanding and perception are crucial components for various applications, including robotics, and autonomous vehicles.  The LiDAR (Light Detection and Ranging) technology contributes to that by providing a three-dimensional and high-resolution representation of the surrounding environment. Specifically, the  4D panoptic LiDAR segmentation\cite{aygun20214d} task which was recently introduced, utilizes the LiDAR data to enable the simultaneous segmentation of objects into semantic classes and the tracking of their movements over time, wishing to unify semantic segmentation, instance segmentation, and tracking into a single framework.



Several existing methods address the 4D panoptic segmentation task, including, 4D-PLS\cite{aygun20214d} and 4D-Stop\cite{kreuzberg2022stop} which employ a tracking-by-detection methodology comprising two primary steps. The first step involves detection, where panoptic segmentation is acquired per frame using neural networks containing semantic segmentation and semantic instance branches. The second step is tracking, where identities are associated with objects over time.

Our examination reveals that the existing works mainly focus on improving the detection step by enhancing the neural network architecture. Surprisingly, the tracking step, crucial for maintaining object identities over time, has received limited attention. Most methods continue to use the naive 4D-PLS tracker \cite{aygun20214d} without modification.

The 4D-PLS tracker adopts a time association approach by matching current time instances with previous time instances. The latter is derived from neural network results computed from a temporal window of $\tau$ consecutive LiDAR scans, encompassing the current time scan. If a match is found, the object retains the same identity as in the previous frames; otherwise, it is considered a new identity for tracking.

However, this approach has notable limitations. Firstly, it relies on short-time matching, proving disadvantageous in scenarios of prolonged occlusion or miss-detection. The failure to leverage tracker motion information further restricts its effectiveness. Secondly, the use of non-adaptive thresholds leads to mismatches,  as matching is only conducted for large-sized object detection. These issues collectively contribute to numerous cases of ID switches, highlighting the need for a more sophisticated and adaptable tracking methodology.

In this work, we address the current limitations of the tracking process by introducing an improved tracking methodology called the \textbf{NextStop} tracker. Drawing inspiration from the well-established tracking method SORT (Simple Online and Real-Time Tracker) \cite{bewley2016simple}, our NextStop tracker employs Kalman filtering for motion estimation and the Hungarian algorithm for associating object detections across frames, while integrating a tracklet state concept to prioritize trackers and detections. Additionally, our NextStop tracker utilizes the collected tracking data to address temporal inconsistencies in the semantic segmentation results. Integrated into 4D-STOP \cite{kreuzberg2022stop}, a leading approach for the 4D panoptic LiDAR segmentation task, NextStop replaces the tracking block and significantly improves the LSTQ metric \cite{aygun20214d} compared to existing methods, as shown in \autoref{tab:sum-results-mots-lstq-identity}. Our NextStop tracker demonstrates superior continuity, reduced ID switches, and earlier tracking initiation compared to alternative methods, with particular benefits for small-sized objects like \textit{person} and \textit{bicyclist}.


%% file: sections/related_work.tex
\section{Related Work}

\PAR{Tracking-by-detection.} 
Tracking-by-detection is a method where objects are first detected in each frame using an object detector, and then tracked across frames by matching these detections using data association techniques. Data association methods are categorized into online (casual) methods, which seek optimal associations between past and current frames, and offline (non-casual) methods, which aim for global optimal associations across the entire sequence including past, current, and future frames.
SORT (Simple Online and Real-Time) \cite{bewley2016simple} by Alex Bewley et al. (2016) is an early online real-time 2D multi-object tracking algorithm that uses Faster-RCNN \cite {ren2015faster} for object detection and Kalman filter \cite{bishop2001introduction} predictions for tracking but does not handle occlusions or re-entry of objects.
AB3DMOT \cite{Weng2020_AB3DMOT} extends the 2D SORT\cite{bewley2016simple} to 3D by incorporating a constant velocity Kalman filter and supports multi-class tracking with similarity metrics such as 3D-IoU and 3D-GIoU \cite{Rezatofighi_2018_CVPR}.

\PAR{4D panoptic segmentation.} The 4D panoptic segmentation task aims to assign distinct instance IDs and semantic labels to every point in a point cloud, expanding panoptic segmentation to include the temporal dimension. The first solution for this task, 4D Panoptic LiDAR Segmentation (4D-PLS) \cite{aygun20214d} by Aygun et al. (2021), treats segmentation and tracking as distinct tasks within a tracking-by-detection framework. It uses a 4D point cloud created from consecutive LiDAR scans as input for an Encoder-Decoder deep learning network that predicts panoptic segmentation per frame. Tracking is performed by matching instances across a temporal window using a cost matrix and the Hungarian algorithm, though it excludes small-sized objects and suffers from issues like frequent ID switches, hard-coded thresholds that aren’t adaptable to all classes, lack of motion estimation, and fast birth which creates creation of false positives. The subsequent 4D-STOP \cite{kreuzberg2022stop} method improves panoptic segmentation  per frame by revising the network architecture and using an instance-centric voting approach but still relies on the same tracking algorithm as 4D-PLS \cite{aygun20214d}, leaving tracking challenges unaddressed.

%% file: sections/method.tex
\input{figures2/method/4D_Panoptic_LiDAR_Segmentation_and_Tracking_pipeline}

\section{Method}
\subsection{Motivation}

LiDAR technology measures ground object distances with laser beams, creating 3D point clouds that face challenges in tracking due to sparse and irregular point distributions. The 2021 4D-PLS \cite{aygun20214d} framework by Aygun et al. addresses these challenges using a two-stage tracking-by-detection approach. It first detects semantic and instance segmentation for each point in the point cloud and then associates these detections over time to ensure consistent object identification.

Building on 4D-PLS\cite{aygun20214d}, subsequent methods like Kreuzberg's 4D-Stop \cite{kreuzberg2022stop} have focused on improving detection while retaining the original tracking methodology\cite{aygun20214d}, which has limitations. The current approach struggles with prolonged occlusions, fixed thresholds leading to identity switches, and limited motion estimation. Additionally, misassignments of segmentation labels, such as confusing different vehicle types, highlight the need for more advanced tracking techniques. This work aims to refine tracking-by-detection methodologies to enhance object perception and identification in LiDAR systems.

\subsection{NextStop Tracker}
we introduce NextStop, our novel tracking mechanism intended to enhance and replace the tracking component of the 4D Panoptic LiDAR Segmentation network, 4D-STOP \cite{kreuzberg2022stop}. As illustrated in \autoref{fig:method_pipline}, NextStop takes as input the panoptic segmentation per frame, which is generated by the pre-trained 4D-STOP \cite{kreuzberg2022stop} neural network. The NextStop framework consists of two primary stages: the bounding box tracker stage and the bounding box to point label stage. Detailed descriptions of these stages will be provided in the following sections.

\input{sections/method_stage1_bounding_box_tracker}

\input{sections/method_stage2_bounding_box_to_points}

%% file: figures2/method/4D_Panoptic_LiDAR_Segmentation_and_Tracking_pipeline

\usetikzlibrary{shadows,arrows,fit,calc,positioning}

\begin{figure}[ht!]
\resizebox{\textwidth}{!}
{
\begin{tikzpicture}

\tikzstyle{RoundRectangleBlock} = [rectangle, rounded corners, minimum width=3cm, minimum height=1.55cm,text centered, draw=black, fill=red!30]
\tikzstyle{arrow} = [thick,->,>=stealth]

\tikzstyle{arrow} = [thick,->,>=stealth]

\pgfdeclarelayer{background}
\pgfdeclarelayer{foreground}
\pgfsetlayers{background,main,foreground}

\node (input)[] {Point Cloud};
\node (4DSTOP_box) [RoundRectangleBlock,right of=input, xshift=3cm,align=center] {4D-STOP net};
\node (tracker_box) [RoundRectangleBlock, right of=4DSTOP_box, xshift=7cm,align=center,fill=green!30] {Stage 1:\\The Bounding-Box Tracker\\\autoref{fig:method_bounding_box_tracker_pipelineB}};


\node (box2label_box) [RoundRectangleBlock, right of=tracker_box, xshift=7cm,yshift=-1.7cm,align=center,fill=blue!30] {Stage 2:\\Bounding box\\to point label \\ \autoref{fig:method_stage2_pipline}};

\node (out_inst)[below of =box2label_box,xshift=0.8cm,yshift = -2.8cm,,align=center] {Instance\\per point};
\node (out_seg)[below of =box2label_box,xshift=-0.8cm,yshift = -3.05cm,align=center] {Segm-\\entation\\per point};

\begin{pgfonlayer}{background}
\node[draw, dotted, rounded corners, inner xsep=1em, inner ysep=1em,left=1em of tracker_box.north west,minimum width=14cm, minimum height=5.9cm,line width=0.7mm, fit=(tracker_box.north west) (box2label_box),fill=yellow!50,fill opacity=0.3] (box) {};
\node[fill=white] at ($(box.south west)+(3,0)$) {\textbf{\large{NextStop Tracker (Ours)}}};
\end{pgfonlayer}

\draw [arrow] (input) -- (4DSTOP_box);

\path (4DSTOP_box.east) -- (4DSTOP_box.north east) coordinate[pos=0.5] (4DSTOP_box_east_top);
\path (tracker_box.west) -- (tracker_box.north west) coordinate[pos=0.5] (tracker_box_west_top);
\draw[arrow]  (4DSTOP_box_east_top) -- node[anchor=east,above,align=left,pos=0.25]{Instance\\per point}(tracker_box_west_top);

\path (4DSTOP_box_east_top) -- (tracker_box_west_top) coordinate[pos=0.8] (mid_instance_arrow);
\path (box2label_box.north) -- (box2label_box.north east) coordinate[pos=0.5] (box2label_box_top_east);

\draw [arrow] (mid_instance_arrow)node[black] {$\bullet$}  -- ++(0cm,1.3cm) -| (box2label_box_top_east);

\path (4DSTOP_box.east) -- (4DSTOP_box.south east) coordinate[pos=0.5] (4DSTOP_box_east_bottom);
\path (tracker_box.west) -- (tracker_box.south west) coordinate[pos=0.5] (tracker_box_west_bottom);
\draw[arrow]  (4DSTOP_box_east_bottom) -- node[anchor=east,below,align=center,pos=0.3]{ Segmentation\\per point}(tracker_box_west_bottom);

\path (4DSTOP_box_east_bottom) -- (tracker_box_west_bottom) coordinate[pos=0.8] (mid_segmentation_arrow);
\draw [arrow] (mid_segmentation_arrow)node[black]{$\bullet$} |-
(box2label_box.west);

\path (box2label_box.north) -- (box2label_box.north west) coordinate[pos=0.4] (box2label_box_top_west);

\draw[arrow]  (tracker_box.east)node[anchor=east,xshift=4.2cm,yshift=0.2cm]{Bounding Box Tracker}-|(box2label_box_top_west);

\path (box2label_box.south)--(box2label_box.south west) coordinate[pos=0.52] (out1);
\draw[arrow] (out1) -- (out_seg);

\path (box2label_box.south)--(box2label_box.south east) coordinate[pos=0.5] (out2);
\draw[arrow] (out2) -- (out_inst);

\end{tikzpicture}
}
\caption{NextStop Block Diagram} \label{fig:method_pipline}
\end{figure}
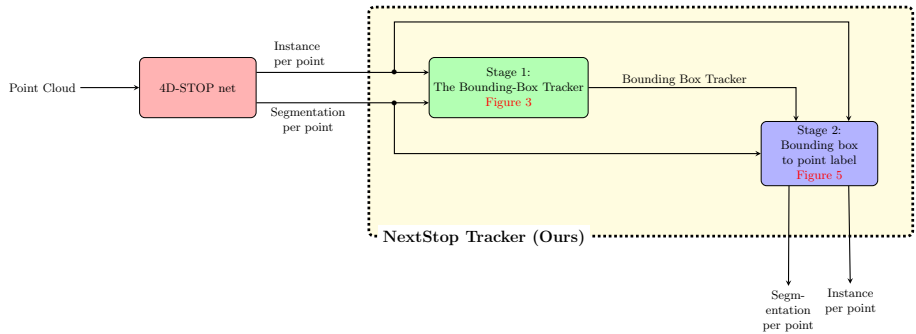

%% file: sections/method_stage1_bounding_box_tracker.tex
\subsection{The Bounding-Box Tracker (Stage 1)}
\label{sec:method_stage1_overview}
In the tracking-by-detection approach, once panoptic segmentation provides the object detections, the next step is to link these detected objects across frames to establish their trajectories over time. Inspired by the SORT \cite{bewley2016simple} approach, we used the AB3DMOT\cite{Weng2020_AB3DMOT} tracker as a base framework to create the \textit{NextStop} box tracker framework which is shown in \autoref{fig:method_bounding_box_tracker_pipelineB}.

\input{figures2/method/stage1/bb_track_flow_tikzB}

\subsubsection{Detection}
\label{sec:method_stage1_Detection}
At each frame \textit{t} an object detection network detects objects and produces a set of detections. Our study utilized 4D-STOP network results \cite{kreuzberg2022stop}. These panoptic per-point results were converted to bounding box detections. Each bounding box is represented by its center-point \textit{($c_x$, $c_y$, $c_z$)}, orientation angle \textit{$\theta$}, dimension measurement  \textit{($l$, $w$, $h$)}, and score \textit{s} that corresponds to the highest score of point associated with the object. Due to the lack of orientation information in the SemanticKITTI database, $\theta$ was always zero. The bounding box detections are then divided into \textit{low score detection boxes} and \textit{high score detection boxes}, for use at later stages.

\subsubsection{Motion-Based Prediction} 
\label{sec:method_stage1_Motion-Based Prediction}
A motion model is a mathematical representation that forecasts the evolving state of an object over time, covering various attributes like position and velocity. The predicted state can then be compared to the value obtained through detection and subsequently adjusted based on the detections observed. In our framework, we used the Kalman filter \cite{bishop2001introduction} with a constant-velocity model for motion-based prediction.
For each tracked object, we maintained a state vector of ten-dimensional:
${x=\left[ c_x, c_y, c_z, \theta, l, w, h, v_x, v_y, v_z \right]^T}$.
the first 7 elements ${(c_x, c_y, c_z, \theta, l, w, h)}$
represent the 3D bounding box: centroid, orientation, and dimension measurements, where the last elements ${(v_x, v_y, v_z)}$ represent the 3D bounding box velocity.

We utilized the prediction equation of the Kalman filter to forecast the state of the tracker. Subsequently, if the tracker was associated with detection, we utilized the correction equation of the Kalman filter to modify its state. The observation vector in this scenario is a seven-dimensional vector represented as ${z=\left[ c_x, c_y, c_z, \theta, l,w, h \right]^T}$, which encompasses the centroid, orientation, and dimension measurements of the 3D detection bounding box. To address the absence of orientation data in the SemanticKITTI database, we reduced the influence of the orientation component in both the state and measurement vectors by specifying the Kalman matrices in a particular manner. 
Furthermore, we set specific parameters for certain categories of classes based on their unique characteristics. For more detailed information on the implementation, check \autoref{sec:kalman}.

\subsubsection{Tracklets State}
\label{sec:method_stage1_Tracklets States}
A tracklet can exist in either a \textit{candidate} state or an \textit{active} state at any given time. Generally, the \textit{active} state refers to tracklets with a high level of credibility, while the \textit{candidate} state is for those with some uncertainty, considered for termination or transition to the \textit{active} state. Tracklets in the \textit{candidate} state aren't included in final tracking results and remain concealed. Upon creation of a new tracklet, it immediately enters the \textit{candidate} state as its reliability hasn't been established yet, which is refined based on motion estimation and data association. Transition from the \textit{candidate} to the \textit{active} state occurs when there are at least \textit{hits} of frames with matched detection and fewer than \textit{max age} frames with no associated detection. Transition from \textit{candidate} state to termination happens when there were more than \textit{death age} frames without matched detection, where \textit{death age} significantly exceeds \textit{max age}. Transition from \textit{active} to \textit{candidate} state occurs when there were more than \textit{max age} frames without matched detection.

\subsubsection{Data Association}
\label{sec:method_stage1_Data Association}
The data association model addresses the \textit{bipartite graph} problem of finding matches between detection and tracklets. Its outcomes consist of the following: (i) matched pairs of tracklets to detection; (ii) unmatched tracklets;(iii) unmatched detections;
Assuming that the current frame, denoted as \textit{k}, has M distinct detections, which are represented by the set $D_k = \{d_1, d_2, ..., d_M\}$, and N distinct tracklet predictions, represented by the set $T_k = \{t_1, t_2, ..., t_N\}$, we begin by constructing the affinity matrix A. This matrix has dimensions MxN and each element is filled using the 3D Distance-IoU (DIoU) similarity metric \cite{zheng2020diou}: $[A]_{i,j} = DIoU(d_i,t_j)\Hquad \textrm{for every} \Hquad i \in M \Hquad \textrm{and} \Hquad j \in N$. Then we use the Hungarian method \cite{kuhn1955hungarian} to find matches, and then we remove pairings that have a similarity value below the specified \textit{threshold}.

\subsubsection{Class Majority}
Upon associating a tracker with a detection, we not only applied the Kalman correction equation but also leveraged the detection class information to modify the tracker's class ID. In our observations, the segmentation output of the \textit{Things} class category from the 4D-STOP \cite{aygun20214d} network sometimes exhibited temporal inconsistency. Therefore, we decided to assign the tracker's class ID based on the most common class type among the detections associated to that tracker

\subsubsection{Prioritization}
\label{sec:method_stage1_Prioritization}
Prioritization involves implementing policies and practices to benefit specific targets based on defined criteria or preferences. In our efforts to enhance data association outcomes, we integrated two prioritization mechanisms.

Firstly, inspired by the methodology outlined in ByteTrack’s research \cite{zhang2022bytetrack}, we divided the detection boxes into two groups based on their scores: high score and low score. Initially, we pair high score detection boxes with tracklets. However, certain tracklets may remain unmatched if they fail to align with a suitable high score detection box. In such cases, we proceed to associate the low score detection boxes with these unmatched tracklets. Secondly, we incorporated the tracklet state concept. Tracklets labeled as "active state" are are often seen as more trustworthy than those in the "candidate state". Consequently, they are given priority in the data association process.

%% file: figures2/method/stage1/bb_track_flow_tikzB.tex
\usetikzlibrary{fit,calc,shapes.geometric}

\begin{figure}[ht!]
\resizebox{\textwidth}{!}
{
\begin{tikzpicture}

\tikzstyle{txtBlock} = [inner sep=0pt,text centered]
\tikzstyle{RedRectangleBlock} = [rectangle, rounded corners, minimum width=1cm, minimum height=1cm,text centered, draw=black, fill=red!30]
\tikzstyle{GreenRectangleBlock} = [rectangle, rounded corners, minimum width=1cm, minimum height=1cm,text centered, draw=black, fill=green!30]
\tikzstyle{YellowRectangleBlock} = [rectangle, rounded corners, minimum width=4cm, minimum height=1cm,,text centered, draw=black, fill=yellow!30]
\tikzstyle{BlueRectangleBlock} = [rectangle, rounded corners, minimum width=1cm, minimum height=1cm,,text centered, draw=black, fill=blue!30]
\tikzstyle{PurpleRectangleBlock} = [rectangle, rounded corners, minimum width=1cm, minimum height=1cm,,text centered, draw=black, fill=purple!30]

\node (association1)[RedRectangleBlock,align=center,xshift= 3cm] {DIoU Match1};

\node (unmatch_tracks1)[YellowRectangleBlock,right of =association1,align=center,xshift= 3cm,yshift= -1.5cm] {Unmatch Tracks};

\node (matched1)[YellowRectangleBlock,right of =association1,align=center,xshift= 3cm] {Match Tracks};

\node (unmatch_detection1)[YellowRectangleBlock,right of =association1,align=center,xshift= 3cm,,yshift= 1.5cm] {Unmatch Detection};

\node (association2)[RedRectangleBlock,below of =unmatch_tracks1,align=center,yshift= -1cm] {DIoU Match2};

\node (unmatch_tracks2)[YellowRectangleBlock,right of =association2,align=center,xshift= 4cm,yshift= -1.5cm] {Unmatch Tracks};

\node (matched2)[YellowRectangleBlock,right of =association2,align=center,xshift= 4cm] {Match Tracks};

\node (unmatch_detection2)[YellowRectangleBlock,right of =association2,align=center,xshift= 4cm,yshift= 1.5cm] {Unmatch Detection};

\node (update)[PurpleRectangleBlock,right of =matched2,align=center,xshift= 5cm] {Kalman Update};

\node (predict)[PurpleRectangleBlock,below of =association1,align=center,yshift= -5cm] {Kalman Predict};

\node[draw,  ultra thick,dotted,line width=3pt, rounded corners,align=center, fit=(association1)(association2)(update)(unmatch_detection1)(unmatch_tracks1)(predict),label={[align=center] above:Base Block }](box1) {};

\node (in_detection1)[txtBlock,fill=red!80, left of =association1,align=center,xshift=-2cm] {IN1};
\node (in_detection1_txt)[txtBlock, below of =in_detection1,align=center,yshift=0.5cm,xshift=-0.2cm] {(Detection High)};

\node (in_detection2)[txtBlock,fill=red!80,left of =association2,align=center,xshift=-6cm] {IN2};
\node (in_detection2_txt)[txtBlock, below of =in_detection2,align=center,yshift=0.5cm,xshift=-0.2cm] {(Detection Low)};

\node (in_track)[txtBlock,fill=red!80,left of =predict,align=center,xshift=-2cm] {IN3};
\node (in_track_txt)[txtBlock, below of =in_track,align=center,yshift=0.5cm,xshift=-0.1cm] {(IN Tracklets)};

\node (out_unmatch_det1)[txtBlock,fill=yellow!, right of =unmatch_detection1,align=center,xshift=13cm] {OUT1};
\node (out_unmatch_det1_txt)[txtBlock, below of =out_unmatch_det1,align=center,yshift=0.5cm,xshift=1.1cm] {(Unmatch Detection High)};

\node (out_unmatch_det2)[txtBlock,fill=yellow!80,right of =unmatch_detection2,align=center,xshift=8cm] {OUT2};
\node (out_unmatch_det2_txt)[txtBlock, below of =out_unmatch_det2,align=center,yshift=0.5cm,xshift=1.1cm] {(Unmatch Detection Low)};

\node (out_match)[txtBlock,fill=yellow!, below of =update,align=center,yshift=-3cm] {OUT3};
\node (out_match_txt)[txtBlock, below of =out_match,align=center,yshift=0.5cm,xshift=0cm] {(OUT Tracklets)};

\node (out_unmatch_track)[txtBlock,fill=yellow!80,below of =unmatch_tracks2,align=center,yshift=-1.5cm] {OUT4};
\node (out_unmatch_track_txt)[txtBlock, below of =out_unmatch_track,align=center,yshift=0.5cm,xshift=0cm] {(Unmatch Tracklets)};



\draw[-, ultra thick,blue!50](association1.east)--(unmatch_tracks1.west);
\draw[-, ultra thick,blue!50](association1.east)--(unmatch_detection1.west);
\draw[-, ultra thick,blue!50](association1.east)--(matched1.west);

\draw[->, ultra thick](unmatch_tracks1.south)--(association2.north);

\draw[->, ultra thick](predict.north)--($(association1.south)+(0cm,-3.5cm)$)arc(-90:90:0.5)--(association1.south);

\draw[-, ultra thick,blue!50](association2.east)--(unmatch_tracks2.west);
\draw[-, ultra thick,blue!50](association2.east)--(unmatch_detection2.west);
\draw[-, ultra thick,blue!50](association2.east)--(matched2.west);

\draw[->, ultra thick](matched2.east)--(update.west);
\draw[->, ultra thick](matched1.east)-|(update.north);

\draw[->, ultra thick](in_detection1.east)--(association1.west);

\draw[->, ultra thick](in_detection2.east)--(association2.west);

\draw[->, ultra thick](in_track.east)--(predict.west);

\draw[->, ultra thick](unmatch_detection1.east)--(out_unmatch_det1.west);

\draw[->, ultra thick](unmatch_detection2.east)--(out_unmatch_det2.west);

\draw[->, ultra thick](unmatch_tracks2.south)--(out_unmatch_track.north);

\draw[->, ultra thick](update.south)--(out_match.north);

\end{tikzpicture}
}
\caption{Stage 1: Base Block Diagram of the Bounding Box Tracker}
\label{fig:method_bounding_box_tracker_building block}
\end{figure}
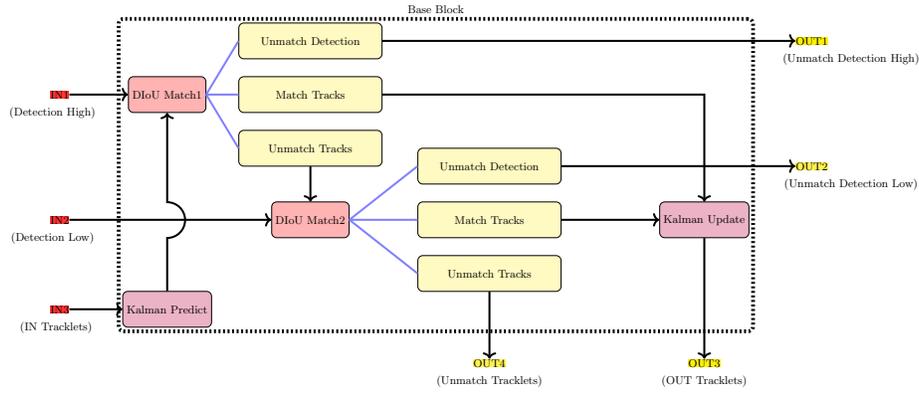

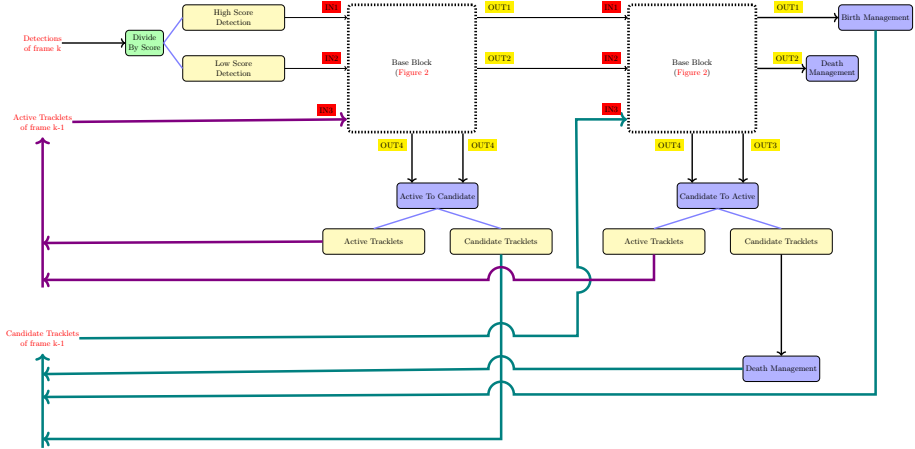
\begin{figure}[ht!]
\resizebox{\textwidth}{!}
{
{

\begin{tikzpicture}

\tikzstyle{txtBlock} = [inner sep=0pt,text centered]
\tikzstyle{RedRectangleBlock} = [rectangle, rounded corners, minimum width=1cm, minimum height=1cm,text centered, draw=black, fill=red!30]
\tikzstyle{GreenRectangleBlock} = [rectangle, rounded corners, minimum width=1cm, minimum height=1cm,text centered, draw=black, fill=green!30]
\tikzstyle{YellowRectangleBlock} = [rectangle, rounded corners, minimum width=4cm, minimum height=1cm,,text centered, draw=black, fill=yellow!30]
\tikzstyle{BlueRectangleBlock} = [rectangle, rounded corners, minimum width=1cm, minimum height=1cm,,text centered, draw=black, fill=blue!30]
\tikzstyle{PurpleRectangleBlock} = [rectangle, rounded corners, minimum width=1cm, minimum height=1cm,,text centered, draw=black, fill=purple!30]
\tikzstyle{BaseBlock} = [ultra thick,dotted,line width=3pt, rounded corners,align=center, minimum width=5cm, minimum height=5cm,,text centered, draw=black]

\node (detection)[txtBlock,align=center,text=red] {Detections\\of frame k};

\node (divide1)[GreenRectangleBlock,right of =detection,align=center,xshift= 3cm] {Divide\\By Score};

\node (highscore_detection)[YellowRectangleBlock,right of =divide1,align=center,xshift= 2.5cm,yshift=1cm] {High Score\\Detection};

\node (lowscore_detection)[YellowRectangleBlock,right of =divide1,align=center,xshift= 2.5cm,yshift=-1cm] {Low Score\\Detection};

\node (active_block)[BaseBlock,right of = highscore_detection,align=center,xshift= 6cm,yshift=-2cm] {Base Block\\(\autoref{fig:method_bounding_box_tracker_building block}};

\node (candidate_block)[BaseBlock,right of = active_block,align=center,xshift= 10cm] {Base Block \\ (\autoref{fig:method_bounding_box_tracker_building block})};

\node (active2candidate)[BlueRectangleBlock,below of =active_block,align=center,xshift= 1cm,yshift= -4cm] {Active To Candidate};

\node (turned2candidate)[YellowRectangleBlock,below of =active2candidate,align=center,xshift= 2.5cm,yshift=-0.8cm] {Candidate Tracklets};

\node (remain_active)[YellowRectangleBlock,below of =active2candidate,align=center,xshift= -2.5cm,yshift=-0.8cm] {Active Tracklets};

\node (track)[txtBlock,align=center,text=red, minimum height=1.2cm,
below of=detection,yshift=-2.1cm] {Active Tracklets \\of frame k-1};



\node (c_track)[txtBlock,align=center,text=red, minimum height=1.2cm,
below of=track,xshift=0cm,yshift=-7.5cm] {Candidate Tracklets \\of frame k-1};

\node (birth_management)[BlueRectangleBlock,right of =candidate_block,align=center,xshift= 6.2cm,yshift= 2cm] {Birth Management};

\node (X)[BlueRectangleBlock,right of =candidate_block,align=center,xshift= 4.5cm] {Death\\ Management};

\node (candidate2active)[BlueRectangleBlock,below of =candidate_block,align=center,xshift= 1cm,yshift= -4cm] {Candidate To Active};

\node (turned2active)[YellowRectangleBlock,below of =candidate2active,align=center,xshift= -2.5cm,yshift=-0.8cm] {Active Tracklets};

\node (remain_candidate)[YellowRectangleBlock,below of =candidate2active,align=center,xshift= 2.5cm,yshift=-0.8cm] {Candidate Tracklets};

\node (death_management)[BlueRectangleBlock,below of =remain_candidate,align=center,yshift= -4cm] {Death Management};;


\draw[->, ultra thick](detection.east)--(divide1);
\draw[-, ultra thick,blue!50](divide1.east)--(highscore_detection.west);
\draw[-, ultra thick,blue!50](divide1.east)--(lowscore_detection.west);

\draw[->] (highscore_detection.east)--node[left,xshift=1cm,yshift=0.4cm,align=center,fill=red]{IN1}($(active_block.west)+(0cm,2cm)$);

\draw[->] (lowscore_detection.east)--node[left,xshift=1cm,yshift=0.4cm,align=center,fill=red]{IN2}(active_block.west);

\draw[->,draw=violet,line width=1mm] (track.east)--node[left,xshift=5cm,yshift=0.4cm,align=center,fill=red]{IN3}($(active_block.west)+(0cm,-2cm)$);


\draw[->] ($(active_block.east)+(0cm,2cm)$)--node[left,xshift=2.7cm,yshift=0.4cm,align=center,fill=red]{IN1}($(candidate_block.west)+(0cm,2cm)$);

\draw[->] ($(active_block.east)+(0cm,2cm)$)--node[left,xshift=-1.5cm,yshift=0.4cm,align=center,fill=yellow]{OUT1}($(candidate_block.west)+(0cm,2cm)$);

\draw[->] (active_block.east)--node[left,xshift=2.7cm,yshift=0.4cm,align=center,fill=red]{IN2}(candidate_block.west);

\draw[->] (active_block.east)--node[left,xshift=-1.5cm,yshift=0.4cm,align=center,fill=yellow]{OUT2}(candidate_block.west);

\draw[->, ultra thick]($(active_block.south)+(2cm,0cm)$)--node[right,xshift=0.2cm,yshift=0.5cm,align=center,fill=yellow]{OUT4}($(active2candidate.north)+(1cm,0cm)$);

\draw[->, ultra thick]($(active_block.south)+(0cm,0cm)$)--node[left,xshift=-0.2cm,yshift=0.5cm,align=center,fill=yellow]{OUT4}($(active2candidate.north)+(-1cm,0cm)$);

\draw[-, ultra thick,blue!50](active2candidate.south)--(turned2candidate.north);
\draw[-, ultra thick,blue!50](active2candidate.south)--(remain_active.north);

\draw[->, ultra thick,draw=violet,line width=1mm] (remain_active.west)--($(track.south)+(0cm,-4.15cm)$);

\draw[->, ultra thick,draw=violet,line width=1mm] ($(track.south)+(0cm,-5.9cm)$)--(track);



\draw[->, ultra thick,draw=teal,line width=1mm](c_track.east)--
($(turned2candidate.south)+(-0.5cm,-3.2cm)$)arc(180:0:0.5)
-|($(turned2candidate.south)+(3cm,-1.5cm)$)arc(-90:90:0.5)--
($(candidate_block.west)+(-2cm,-2cm)$)--
node[left,xshift=0.8cm,yshift=0.4cm,align=center,fill=red]{IN3}($(candidate_block.west)+(0cm,-2cm)$);

\draw[->, ultra thick]($(candidate_block.east)+(0cm,2cm)$)--node[left,xshift=0.2cm,yshift=0.4cm,align=center,fill=yellow]{OUT1}(birth_management.west);

\draw[->, ultra thick](candidate_block.east)--
node[left,xshift=0.8cm,yshift=0.4cm,align=center,fill=yellow]{OUT2}(X.west);



\draw[->, ultra thick]($(candidate_block.south)+(2cm,0cm)$)--node[right,xshift=0.3cm,yshift=0.5cm,align=center,fill=yellow]{OUT3}($(candidate2active.north)+(1cm,0cm)$);

\draw[->, ultra thick]($(candidate_block.south)+(0cm,0cm)$)--node[left,xshift=-0.3cm,yshift=0.5cm,align=center,fill=yellow]{OUT4}($(candidate2active.north)+(-1cm,0cm)$);

\draw[-, ultra thick,blue!50](candidate2active.south)--(turned2active.north);
\draw[-, ultra thick,blue!50](candidate2active.south)--(remain_candidate.north);

\draw[->, ultra thick](remain_candidate.south)--(death_management.north);

\draw[->, ultra thick,,draw=teal,line width=1mm] (birth_management.south)|-($(turned2candidate.south)+(0.5cm,-5.5cm)$)arc(0:180:0.5)--($(c_track.south)+(0cm,-1.7cm)$);

\draw[->, ultra thick,,draw=teal,line width=1mm] (death_management.west)--($(turned2candidate.south)+(0.5cm,-4.5cm)$)arc(0:180:0.5)--($(c_track.south)+(0cm,-0.8cm)$);

\draw[->, ultra thick,,draw=teal,line width=1mm] ($(c_track.south)+(0cm,-3.7cm)$)--(c_track);


\draw[->, ultra thick,draw=teal,line width=1mm] (turned2candidate.south)|-($(c_track.south)+(0cm,-3.35cm)$);

\draw[->, ultra thick,draw=violet,line width=1mm] (turned2active.south)|-($(turned2candidate.south)+(0.5cm,-1cm)$)arc(0:180:0.5)--($(track.south)+(0cm,-5.6cm)$);

\end{tikzpicture}
}
}
\caption{Stage 1: Bounding Box Tracker Complete Block Diagram}
\label{fig:method_bounding_box_tracker_pipelineB}
\end{figure}

%% file: sections/method_stage2_bounding_box_to_points.tex
\subsection{Bounding-box to Point Label (Stage2)}
\label{sec:method_stage2_overview}

The initial stage, known as the Bounding-Box Tracker (outlined in \autoref{sec:method_stage1_overview}), yields tracking results in the form of bounding box trackers. However, for the task of \textit{4D Panoptic LiDAR Segmentation and Tracking}, the expected tracking results are in a "per-point" format. This implies that each point belonging to a tracked object should possess a label exclusively assigned to that object and unique over time. To accommodate this requirement, Stage 2 was introduced, where we provide a tracking label per point derived from the bounding box track label. Refer to \autoref{fig:method_stage2_simple_example} for an illustration of this process.

\input{figures2/method/stage2/stage2_simple_example}

\input{figures2/method/stage2/stage2_pipline}

To generate per-point tracking results, we leverage the outputs from both the 4D-STOP \cite{kreuzberg2022stop} and the results from our bounding box tracker as inputs, as depicted in \autoref{fig:method_stage2_pipline}:  First, each bounding box tracker is associated with the points corresponding to the object it tracks, while managing any overlaps between boxes that may occur. Then, a distinct track ID label is assigned to the associated points, ensuring uniqueness across all classes. Next, instance IDs are allocated to the remaining unassociated points.

\subsubsection{Associating Tracked Bounding Boxes with Points}
\label{sec:method_stage2_Association points to bounding box}

In this block, we are associating points to their corresponding bounding box tracker, as shown in \autoref{fig:method_stage2_getIndToPointsInsideBox_example}. First, the points located within the bounding box are extracted. Not all of these points belong to the object that we are tracking. Some belong to \textit{stuff} class category like \textit{vegetation} and \textit{terrain}, and some may be from a different \textit{Thing} class category near the target. In addition, despite accurate center point prediction, the bounding box tracker often struggles with dimension estimation, leading to points exceeding the box boundaries. The absence of orientation angle information exacerbates the issue, causing certain tracked object points to fall outside the box.

To overcome this, we leverage the 4D-STOP \cite{kreuzberg2022stop} network's initial instance and segmentation results. Once the points within the bounding box are extracted, we eliminate the points that belong to the \textit{stuff} class category. Then, we utilize KDTree to identify the nearest instance to the remaining inner points. The points associated with this bounding box are determined by selecting the union of the inner box \textit{Things} class points and the points that belong to the closest instance.
This guaranteed that the related points are not constrained by a bounding box and belong to the \textit{Things} class category.

\input{figures2/method/stage2/stage2_getIndToPointsInsideBox}

\subsubsection{Managing Overlapping Points}
\label{sec:method_stage2_Handle overlapping points}
Having assigned points to each bounding box, we now define overlap between bounding boxes as the overlap of their associated points. If two boxes share more than three common points, it indicates that they are overlapping. To maintain simplicity and accommodate the overlap, we compute the ratio of the shared points to the total points within each box. The common points will be assigned to the box that has the highest ratio,  indicating that most of the common points belong to it. An example of this process is shown in \autoref{fig:method_stage2_overlap_example}.

\input{figures2/method/stage2/stage2_overlap_example}

\subsubsection{Ensuring Temporal Consistency in Setting Points with ID}
\label{sec:method_stage2_Fill points with ID consist over time}

This part aims to assign unique instance IDs and semantic labels to each point, so the instance ID is unique spatially and temporally.

We start labeling all points associated with the bounding box tracker. 
Each bounding box tracker already possessed the class information and the track ID as it was set in stage 1 (\ref{sec:method_stage1_overview}).
The class information was used to assign the semantic class to the points associated with each box. However, the track ID could not be used directly because it was not unique across all classes. This resulted in boxes from different classes potentially sharing the same track ID, even if they belonged to different objects. To address this issue we introduce a memory structure named \textit{IDMemory}. This structure serves as a lookup table that links the bounding box track ID and class ID to a unique instance ID label. 
Next, we proceed to label the remaining unlabeled points. We set their class ID and instance ID based on the segmentation and instance results generated by the 4D-STOP \cite{kreuzberg2022stop} network, following an approach similar to that of the 4D-STOP  \cite{kreuzberg2022stop} and 4D-PLS \cite{aygun20214d} trackers. Objects classified as belonging to the \textit{Things} class that were smaller than 25 points had both their class ID and instance ID set to zero. The class ID and instance ID of objects from the \textit{Stuff} or \textit{Things} class of large points size are both assigned to be equal to the class ID they obtained from the 4D-STOP results \cite{kreuzberg2022stop}.

%% file: figures2/method/stage2/stage2_simple_example.tex
\begin{figure}[ht!]
\resizebox{\textwidth}{!}
{
\begin{subfigure}{.25\textwidth}
  \centering
  \tikz[remember picture]\node[inner sep=0pt,outer sep=0pt] (a){\includegraphics[width=.7\linewidth]{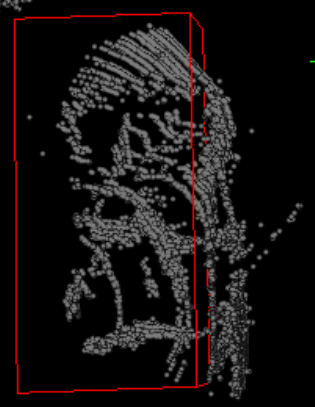}};
  \caption{}
\end{subfigure}%

\begin{subfigure}{.25\textwidth}
  \centering
  \tikz[remember picture]\node[inner sep=0pt,outer sep=0pt] (b){\includegraphics[width=.7\linewidth]{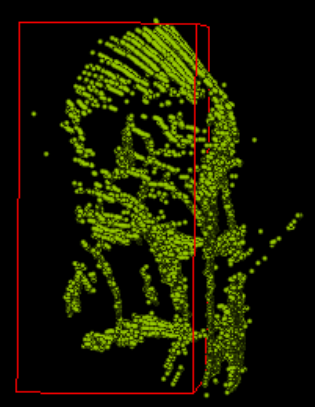}};
  \caption{}
\end{subfigure}

\begin{subfigure}{.25\textwidth}
  \centering
  \tikz[remember picture]\node[inner sep=0pt,outer sep=0pt] (c){\includegraphics[width=.7\linewidth]{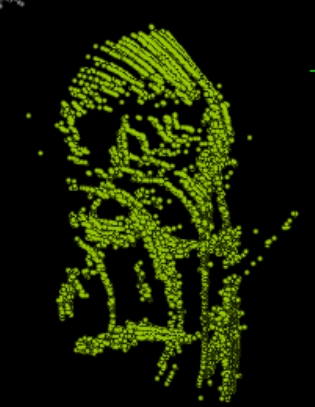}};
  \caption{}
\end{subfigure}

\tikz[remember picture,overlay]\draw[line width=2pt,-stealth,black] ([xshift=2mm]a.east) -- ([xshift=-2mm]b.west)node[midway,above,text=black,font=\LARGE\bfseries\sffamily] {};

\tikz[remember picture,overlay]\draw[line width=2pt,-stealth,black] ([xshift=2mm]b.east) -- ([xshift=-2mm]c.west)node[midway,above,text=black,font=\LARGE\bfseries\sffamily] {};

}
\begin{flushleft}
\small
{
\textbf{(a)} Points belonging to a car object in gray. The tracked bounding box is in \textcolor{red}{\textbf{red}}. \textbf{(b)} Result after the association. Points in \textcolor{green}{\textbf{green}} associated with the \textcolor{red}{\textbf{red}} tracked bounding box. \textbf{(c)} Final results, which do not contain a bounding box.
}
\end{flushleft}

\caption{Stage 2: From Bounding Box to Points with ID} 
\label{fig:method_stage2_simple_example}
\end{figure}

%% file: figures2/method/stage2/stage2_pipline.tex
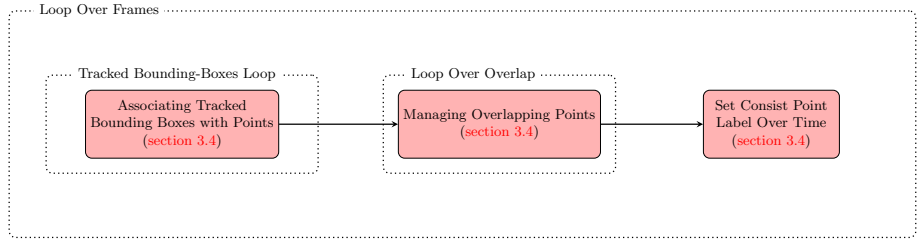
\begin{figure}[ht!]
\resizebox{\textwidth}{!}
{
\begin{tikzpicture}

\tikzstyle{RoundRectangleBlock} = [rectangle, rounded corners, minimum width=3cm, minimum height=1.5cm,text centered, draw=black, fill=red!30]
\tikzstyle{arrow} = [thick,->,>=stealth]
\tikzstyle{arrow} = [thick,->,>=stealth]

\node (ass_box) [RoundRectangleBlock, xshift=3cm,align=center] {Associating Tracked\\ Bounding Boxes with Points\\(\autoref{sec:method_stage2_Association points to bounding box})};
\node (overlap_box) [RoundRectangleBlock, right of=ass_box, xshift=6cm,align=center] {Managing Overlapping Points\\(\autoref{sec:method_stage2_Handle overlapping points})};
\node (consist_label_box) [RoundRectangleBlock, right of=overlap_box, xshift=5cm,align=center] {Set Consist Point\\Label Over Time\\(\autoref{sec:method_stage2_Fill points with ID consist over time})};

\node[draw, thick, dotted, rounded corners, inner xsep=1em, inner ysep=1em,minimum width=20cm, minimum height=5cm, fit=(ass_box.north west) (consist_label_box)] (frame_loop_box) {};
\node[fill=white] at ($(frame_loop_box.north west)+(2,0)$) {Loop Over Frames};

\node[draw, thick, dotted, rounded corners, inner xsep=1em, inner ysep=1em,,minimum width=6cm, fit=(ass_box)] (tracked_loop_box) {};
\node[fill=white] at ($(tracked_loop_box.north west)+(2.9,0)$) {Tracked Bounding-Boxes Loop};

\node[draw, thick, dotted, rounded corners, inner xsep=1em, inner ysep=1em, fit=(overlap_box)] (overlap_loop_box) {};
\node[fill=white] at ($(overlap_loop_box.north west)+(2,0)$) {Loop Over Overlap};

\draw [arrow] (ass_box) -- (overlap_box);
\draw [arrow] (overlap_box) -- (consist_label_box);

\end{tikzpicture}
}
\caption{Stage 2: Bounding-box to Label Per Point Block Diagram} \label{fig:method_stage2_pipline}
\end{figure}

%% file: figures2/method/stage2/stage2_getIndToPointsInsideBox.tex
\begin{figure}[ht!]
\resizebox{\textwidth}{!}
{
\includegraphics[]{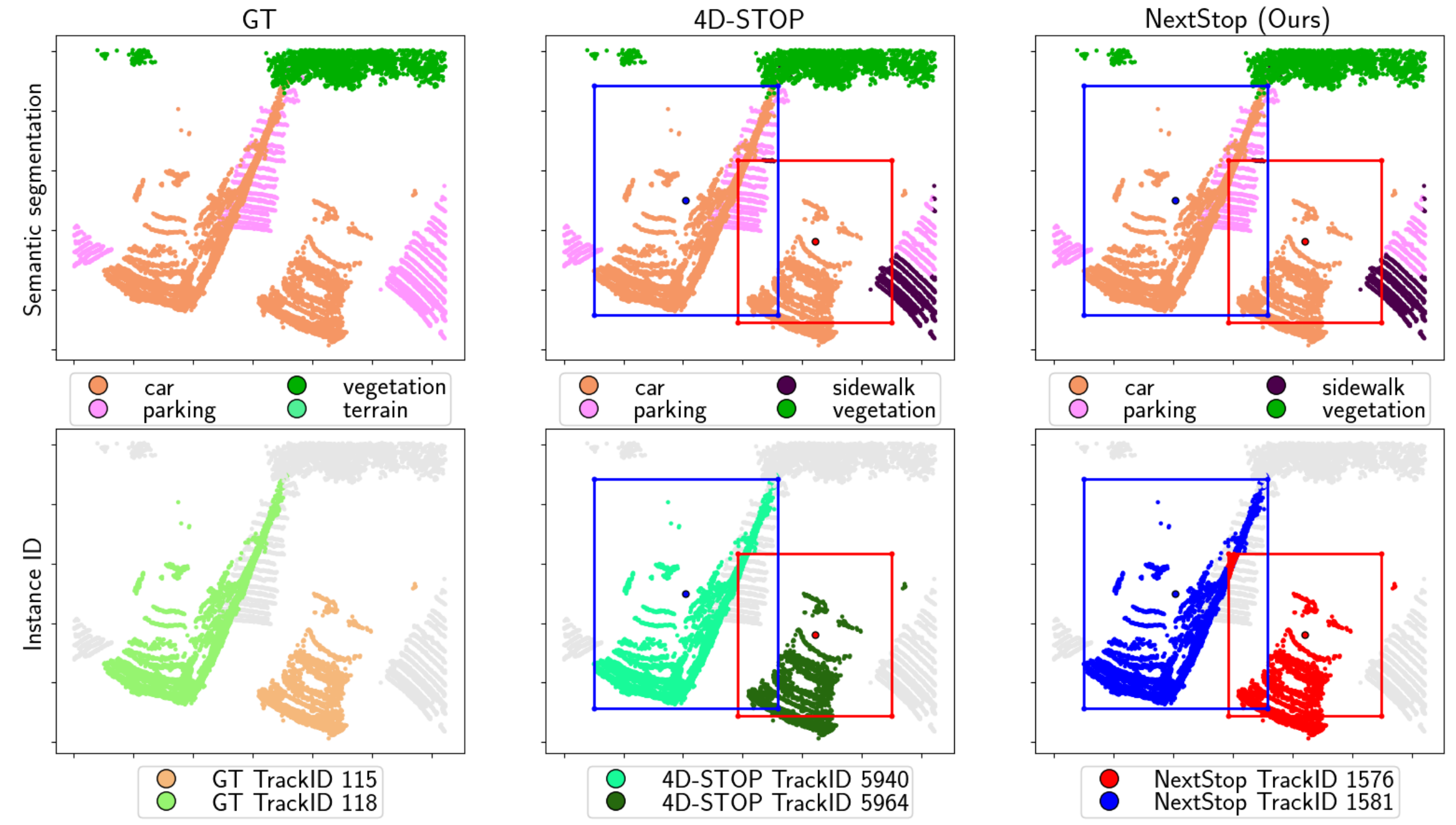}
}
\begin{flushleft}
\small
{
The first row represents semantic segmentation, and the second row represents the instance ID, also known as trackID.
The left column displays the GT data, the middle column shows the results of the 4D-Stop \cite{kreuzberg2022stop} network combined with our stage 1 bounding box tracker results (\ref{sec:method_stage1_overview}), with each box tracker represented by a different color. The right column illustrates the association of points to their bounding box trackers as the output of \ref{sec:method_stage2_Association points to bounding box}. We can observe from the GT data that we have two \textit{car} objects that are surrounded by objects of the \textit{Stuff} class.  The bounding box trackers displayed in the middle column contain points that do not belong to the tracked object, and some points of are outside the box (as shown by the \textcolor{red}{red} bounding box).
By removing the \textit{stuff} class points and employing KD-Tree to locate the nearest 4D-STOP \cite{kreuzberg2022stop} instance to the box's center, we obtain the final points assignment shown in the right column.
}
\end{flushleft}
\caption{Stage 2: Associating Tracked Bounding Boxes with Points} 
\label{fig:method_stage2_getIndToPointsInsideBox_example}
\end{figure}

%% file: figures2/method/stage2/stage2_overlap_example.tex
\begin{figure}[ht!]
\resizebox{\textwidth}{!}
{
\includegraphics[]{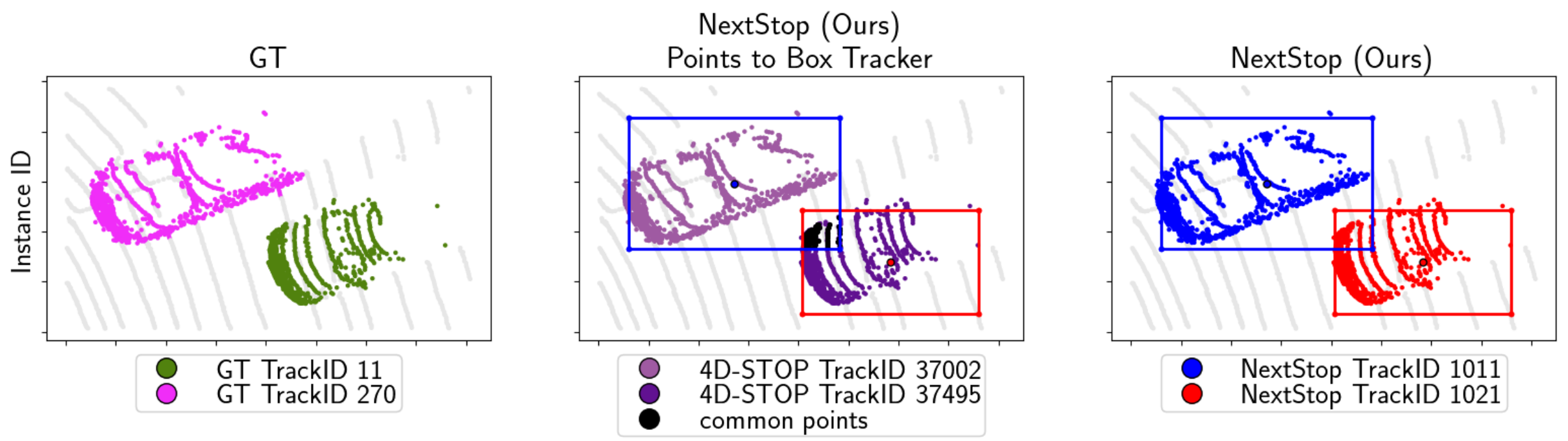}
}

\small
{
\begin{flushleft}
Those graphs all show the instance ID, which is also called the trackID.
The left plot shows the ground truth (GT), the middle displays the result of the previous block \ref{sec:method_stage2_Association points to bounding box}, and the right side demonstrates the final result after addressing the overlap.

This scene features two tracked objects. The bounding box trackers are visible in both the middle and right plots. Both tracking boxes have common points highlighted in black on the middle plot. The overlapping points account for 13.84\% of the points associated with the \textcolor{red}{red} bounding box tracker and only 8.8\% of the \textcolor{blue}{blue} bounding box tracker. This is why they were assigned to the red bounding box tracker, as shown in the plot on the right.
\end{flushleft}
}

\caption{Stage 2: Managing Overlapping Points} 
\label{fig:method_stage2_overlap_example}
\end{figure}

%% file: sections/experiments.tex
\newcommand{\numberone}{\normalsize\ding{172}}%
\newcommand{\numbertwo}{\normalsize\ding{173}}%
\newcommand{\numberthree}{\normalsize\ding{174}}%
\newcommand{\numberfour}{\normalsize\ding{175}}%
\newcommand{\numberfive}{\normalsize\ding{176}}%

\section{Numerical Experiments}

\subsection{Implementations Details}
\label{sec:Implementations Details}

\PAR{The Bounding-Box Tracker:} Discovering the optimal parameter values involved a dual-stage process. The first stage centered on refining the motion estimator, which is based on the Kalman filter. Subsequently, the second stage concentrated on enhancing various other elements including data association, tracker state transitions (from active to candidate and vice versa), and lifespan management. The optimal parameter value varies depending on the class type of the objects (\textit{vehicles}, \textit{bikes}, and \textit{pedestrians}) due to their distinct characteristics, which necessitate different parameter initialization. Specifics regarding the parameters utilized for the Kalman filter can be found in \autoref{sec:kalman}, while details concerning the parameters for the other components are enumerated in \autoref{tab:Implementations-parameters}

\PAR{Bounding-box to Point Label:} In this section, two parameters were specified: bounding box overlap and 'ignore size'. A bounding box overlap was defined when two boxes shared more than three points. Furthermore, an 'ignore size' of 25 was set, indicating that objects categorized as belonging to the \textit{Things} class and smaller than 25 points were assigned a value of zero for both their class ID and instance ID.

\begin{table}[ht!]
\resizebox{\columnwidth}{!}{%
\begin{tabular}{c|c|cccc|cc|c}
 &Detection& \multicolumn{4}{c|}{Data Association} & \multicolumn{2}{c|}{\begin{tabular}[c]{@{}c@{}}State of tracker:\\ active or candidate\end{tabular}} & Kill \\ \cline{2-9} 
 &\begin{tabular}[c]{@{}c@{}}high/low\\ detection threshold\end{tabular}& \begin{tabular}[c]{@{}c@{}}matching \\ metric\end{tabular} & \begin{tabular}[c]{@{}c@{}}matching \\ algorithm\end{tabular}  & \begin{tabular}[c]{@{}c@{}}high detection \\ matching score\end{tabular} & \begin{tabular}[c]{@{}c@{}}low detection\\  matching score\end{tabular} & min hits & max age & death age \\ \cline{2-9} 
Vehicles &0.7& diou & Hungarian  & -0.2 & -0.5 & 2 & 7 & 10 \\
Bikes &0.8& diou & Hungarian & -0.4 & -0.7 & 3 & 4 & 7 \\
Pedestrian & 0.3& diou & Hungarian & -0.4 & -0.7 & 3 & 4 & 7
\end{tabular}%
}
\caption{Implementations Parameters}
\label{tab:Implementations-parameters}
\end{table}

\subsection{Comparing with State-of-the-Art Methods} 
\PAR{Dataset}
We evaluate our method on the SemanticKITTI \cite{behley2019iccv} validation set. The dataset provides point-wise annotations for 28 semantic classes, divided into \textit{Things} category (objects that are capable of moving) and \textit{Stuff} category(objects that are incapable of moving), with unique tracking IDs for the \textit{Things} class.

Following previous studies \cite{aygun20214d,kreuzberg2022stop}, we consolidate the 28 classes into 19 categories, with eight classes belonging to the \textit{Things} category: \textit{car, bicycle, motorcycle, truck, other-vehicle, person, bicyclist, motorcyclist} and 11 classes belonging to the \textit{Stuff} category: \textit{road, parking, sidewalk, other-ground, building, fence, vegetation, trunk, terrain, pole, traffic sign}.

\PAR{Evaluation Metric.}
The designated evaluation metric for the official \textit{4D Panoptic LiDAR Segmentation} is the LSTQ metric, as introduced by Aygun at el. \cite{aygun20214d}. This metric comprises the geometric mean of two scores: the classification score and the association score. In Aygun at el's work, small-size objects, defined as those containing fewer than 50 points, were disregarded solely in the calculation of the association score. Consequently, this approach overlooked not only small and distant objects but also small enclosed objects like pedestrians.

To address this limitation, when evaluating our tracking method, we compared its performance against two variants of the original metric. The first, denoted as $LSTQ_{50}$, followed the protocol of the original metric, while the second, labeled as $LSTQ_1$, included small-size objects in its evaluation without any exclusion criteria.

While both 4D-PLS \cite{aygun20214d} and 4D-StOP \cite{kreuzberg2022stop} were initially assessed using the $LSTQ_{50}$ metric, we opted to evaluate their performance on $LSTQ_1$ as well. This decision stemmed from the fact that the LSTQ metric is not employed as a cost function for optimizing the deep learning networks in either paper. Additionally, the original $LSTQ_{50}$ score only excluded small-sized objects from the association score component, indicating inconsistency in how each score was handled.


\PAR{Results.} We present our results from the validation set. The evaluation results for all classes in the \textit{Things} class category are presented in \autoref{tab:sum-results-mots-lstq-identity}. Results for each class separately can be seen in \autoref{sec:Results_per class}. We conducted a comparative analysis between our NextStop tracker, integrated into the detection network of 4D-STOP \cite{kreuzberg2022stop}, and two existing methods, namely 4D-PLS \cite{aygun20214d} and 4D-STOP \cite{kreuzberg2022stop}.  Both 4D-PLS \cite{aygun20214d} and 4D-STOP \cite{kreuzberg2022stop} utilize identical algorithms for temporal association, but they diverge in their detection network. 

Furthermore, we evaluated the performance of the Tuned Tracking Parameter (TTP) version of both methods. In this version, we adjusted one fixed threshold called \textit{track size}, which belongs to the tracker algorithm of both 4D-PLS\cite{aygun20214d} and 4D-STOP\cite{kreuzberg2022stop}. Our findings indicate that modifying the \textit{track size} parameter leads to more favorable results in comparison to the initial implementations \cite{aygun20214d,kreuzberg2022stop}.
The term \textit{track size} denotes the minimum size of an object at which a consistent tracking ID is maintained over time. The initial version of the work had a \textit{track size} of 50, but in the TTP version, it was decreased to 30. Note that reducing it further did not yield better results as it led to the creation of more false positives in the trackers. More information regarding the \textit{track size} can be found in the source code of 4D-PLS \cite{aygun20214d}.

NextStop achieved a $LSTQ_{50}$ score of 69.65\% for the \textit{Things} class, surpassing the second-place contender, 4D-StOP \cite{kreuzberg2022stop} + TTP, which achieved 68.65\%. While this improvement may seem minor, a significant enhancement was observed in $LSTQ_1$, where NextStop achieved 65.28\%, marking an increase from the second-place 4D-StOP \cite{kreuzberg2022stop} + TTP, which scored 61.71\%. This indicates that our tracking method exhibits a superior performance in tracking small-size objects, as evidenced by the larger growth in $LSTQ_1$. Furthermore, despite our enhancements being primarily focused on improving the tracking component, specifically the $S_{assoc}$ score of the LSTQ metric, the implementation of our tracker also resulted in an enhancement in class segmentation: $S_{cls}$ increased from 64.70\% to 66.76\%.

\input{tables/table}

\PAR{Results per Class}
\label{sec:Results_per class}
In this section, we present the LSTQ\cite{aygun20214d} score results for every class within the \textit{Things} category, which includes cars, bicycles, motorcycles, trucks, other-vehicles, person, and bicyclists.

\input{tables/results_per_class}

From these tables, it is evident that all classes, with the exception of the "bicycle" class, exhibited higher LSTQ scores. Specifically for $LSTQ_1$:
"other-vehicle" increased from 49.14\% to 60.66\%, "person" increased from 44.27\% to 52.66\%, "bicyclist" increased from 39.8\% to 45.5\%, "truck" increased from 72.24\% to 76.05\%, and "car" increased from 86.42\% to 89.64\%.

Deep diving into the scores that contribute to the final $LSTQ_1$  of the five mentioned classes above, we observe an increase of $\pm7\%$ in the association score. Notably, there is a significant improvement for small-sized classes like "person" and "bicyclist," which saw a growth of +10\%. Regarding the classification score, the "other-vehicle" class exhibited a notable increase of +11\%, while the rest of the classes showed minor increases of $\pm 2\%$. These results indicate that our NextStop tracker is especially advantageous for small-sized objects. Moreover, utilizing tracking information to correct temporal inaccuracies in semantic segmentation has significantly benefited the "other-vehicle" class more than the other classes.

In terms of the performance of the "bicycle" class, as observed from the classification component score, S\_\textit{cls}, of the LSTQ metric, this class exhibited the poorest detection score compared to others, with a low score of 33\% for the 4D-PLS detection network\cite{aygun20214d} and 31\% for the 4D-STOP detection network\cite{kreuzberg2022stop}. This may explain why, despite the NextStop tracker surpassing the LSTQ score of 4D-STOP for this class, it did not achieve a higher score than the results obtained by 4D-PLS. Additionally, from the performance of this class compared to the rest of the classes, it is evident that good detection is crucial for effective tracking. 


\subsection{Qualitative Results}
In this section, we demonstrate our tracking method using a few selected examples extracted from the SemanticKITTI\cite{behley2019iccv} validation set, alongside a comparison to the tuned tracking parameter (TTP) version of 4D-STOP \cite{kreuzberg2022stop}, which exhibited better performance compare to the non-tuned one.

\PAR{Successful Examples.} Presenting three successful examples, Figures \autoref{fig:Qualitative_Results_moving-car2_example} and \autoref{fig:Qualitative_Results_moving-person.12_example} showcase the tracking of individual objects. Additionally, in \autoref{fig:Qualitative_Results_solution_motivation_example}, we demonstrate how our NextStop tracker enhances semantic segmentation outcomes, particularly addressing cases of temporal inconsistency.

In \autoref{fig:Qualitative_Results_moving-car2_example}, the tracking of a \textit{Moving Car} is depicted, originating approximately 50 meters away from the LiDAR sensor, presenting a non-trivial tracking challenge due to the object's small size at such distances.

Similarly, in \autoref{fig:Qualitative_Results_moving-person.12_example}, the tracking of a \textit{Moving Person} object is shown. This example also begins far from the LiDAR sensor and poses additional challenges due to the inherently small size of a \textit{Moving Person} object, even when close to the sensor.

These examples highlight that utilizing tracking information from the NextStop tracker aids in improving time-inconsistent semantic class results. Furthermore, our NextStop tracker achieves fewer ID switches and initiates tracking earlier, particularly enhancing trajectory coverage for distant objects.

\PAR{Examining a Challenging Case Study.} In Figure \ref{fig:Qualitative_Results_gt_class_252_gt_id_22_example}, we illustrate the trajectory of a single-tracked object classified as a \textit{moving-car}. During its movement, the object experienced partial occlusion. Despite this challenge, our NextStop tracker demonstrated improvement: it effectively connected the identity observed during occlusion with the post-occlusion identity, leading to fewer ID switches compared to 4D-STOP. However, 4D-STOP initiated tracking of the object earlier during the occlusion, presenting a disadvantage for our NextStop tracking performance during those frames.

\input{figures2/Qualitative_Results/Successes/moving-car2}
\input{figures2/Qualitative_Results/Successes/moving-person}
\clearpage
\input{figures2/Qualitative_Results/Successes/solution_motivation}


\input{figures2/Qualitative_Results/Successes/moving-car}

%% file: tables/table.tex
\clearpage
\begin{table}[ht!]
\resizebox{\columnwidth}{!}{%

\begin{tabular}{lcccccc}
\multicolumn{7}{c}{\textbf{Things}} \\ \hline\hline
\multicolumn{1}{c}{}          & \multicolumn{3}{c}{LSTQ${_1}$}    & \multicolumn{3}{c}{LSTQ${_{50}}$} \\ \hline
\multicolumn{1}{c|}{Method} &
  LSTQ$\uparrow$ &
  \begin{tabular}[c]{@{}c@{}}S\_$assoc$$\uparrow$\end{tabular} &
  \multicolumn{1}{c|}{S\_$cls$$\uparrow$} &
  LSTQ$\uparrow$ &
  \begin{tabular}[c]{@{}c@{}}S\_$assoc$$\uparrow$\end{tabular} &
  S\_$cls$$\uparrow$ \\ \hline
\multicolumn{1}{l|}{4D-PLS$\>\>\>\>\>$\cite{aygun20214d}}&51.27&49.56&\multicolumn{1}{c|}{53.04}&57.04&61.35&53.04\\
\multicolumn{1}{l|}{4D-PLS$\>\>\>\>\>$\cite{aygun20214d} + TTP}&51.76&50.52&\multicolumn{1}{c|}{53.04}&56.97&61.20&53.04  \\
\multicolumn{1}{l|}{4D-StOP \cite{kreuzberg2022stop}}&60.56&56.70&\multicolumn{1}{c|}{64.70}&67.34&70.10&64.70\\
\multicolumn{1}{l|}{4D-StOP \cite{kreuzberg2022stop} + TTP}&61.71&58.86&\multicolumn{1}{c|}{64.70}&68.22&71.95&64.70\\ \hline
\multicolumn{1}{l|}{\textbf{NextStop (Ours)}}&\textbf{  {65.28}}&\textbf{  {63.84}}&\multicolumn{1}{c|}{\textbf{  {66.76}}}&{\textbf{  {69.65}}}&{\textbf{  {72.68}}}&{\textbf{  {66.76}} }\\ \hline
\smallskip
TTP - \textit{Tuned Tracking Parameter}
\end{tabular}%
}
\caption{Scores of \textit{Things} Classes on The SemanticKITTI Validation Set}

\label{tab:sum-results-mots-lstq-identity}

\end{table}


%% file: tables/results_per_class.tex
\label{appendix:Appendix 1}

\newcommand{\strClass}{Car}
\begin{table}[ht!]
\resizebox{\columnwidth}{!}{%
\begin{tabular}{lccccccccc}
\multicolumn{7}{c}{\textbf{\strClass}} \\ \hline\hline
\multicolumn{1}{c}{}   & \multicolumn{3}{c}{LSTQ${_1}$}    & \multicolumn{3}{c}{LSTQ${_{50}}$} \\ \hline
\multicolumn{1}{c|}{Method} &
  LSTQ$\uparrow$ &
  \begin{tabular}[c]{@{}c@{}}S\_$assoc$$\uparrow$\end{tabular} &
  \multicolumn{1}{c|}{S\_$cls$$\uparrow$} &
  LSTQ$\uparrow$ &
  \begin{tabular}[c]{@{}c@{}}S\_$assoc$$\uparrow$\end{tabular} &
  S\_$cls$$\uparrow$ \\ \hline
\multicolumn{1}{l|}{4D-PLS$\>\>\>\>\>$\cite{aygun20214d}}&  80.2& 67 & \multicolumn{1}{c|}{96} &   85.97     &  77      &     96   \\
\multicolumn{1}{l|}{4D-PLS$\>\>\>\>\>$\cite{aygun20214d} + TTP}& 80.8 & 68 & \multicolumn{1}{c|}{96} &     85.97   &     77   &     96   \\
\multicolumn{1}{l|}{4D-StOP \cite{kreuzberg2022stop}}& 85.3 &  75  & \multicolumn{1}{c|}{97} &   91.86     &   87     &     97   \\
\multicolumn{1}{l|}{4D-StOP \cite{kreuzberg2022stop} + TTP} & 86.42 & 77 & \multicolumn{1}{c|}{97} &    \textbf{ {92.39}}    &    \textbf{ {88}}    &    97    \\ \hline
\multicolumn{1}{l|}{\textbf{NextStop (Ours)}}& \textbf{ {89.64}} & \textbf{ {82}} & \multicolumn{1}{c|}{\textbf{ {98}}} &     92.33   &   87    &   \textbf{ {98}}     \\ \hline
\smallskip
TTP - \textit{Tuned Tracking Parameter}
\end{tabular}%
}
\caption{Scores of \textit{\strClass} class on The SemanticKITTI Validation Set}
\label{tab:\strClass-mots-lstq-identity}
\end{table}


\renewcommand{\strClass}{Bicycle}

\begin{table}[ht!]
\resizebox{\columnwidth}{!}{%
\begin{tabular}{lccccccccc}
\multicolumn{7}{c}{\textbf{\strClass}} \\ \hline\hline
\multicolumn{1}{c}{}    & \multicolumn{3}{c}{LSTQ${_1}$}    & \multicolumn{3}{c}{LSTQ${_{50}}$} \\ \hline
\multicolumn{1}{c|}{Method} &
  LSTQ$\uparrow$ &
  \begin{tabular}[c]{@{}c@{}}S\_$assoc$$\uparrow$\end{tabular} &
  \multicolumn{1}{c|}{S\_$cls$$\uparrow$} &
  LSTQ$\uparrow$ &
  \begin{tabular}[c]{@{}c@{}}S\_$assoc$$\uparrow$\end{tabular} &
  S\_$cls$$\uparrow$ \\ \hline
\multicolumn{1}{l|}{4D-PLS$\>\>\>\>\>$\cite{aygun20214d}}& 16.24&8&\multicolumn{1}{c|}{\textbf{ {33}}}& 20.71&13&\textbf{ {33}}  \\
\multicolumn{1}{l|}{4D-PLS$\>\>\>\>\>$\cite{aygun20214d} + TTP}&\textbf{ {18.16}}&\textbf{ {10}}&\multicolumn{1}{c|}{\textbf{ {33}}}& \textbf{ {21.5}}&\textbf{ {14}}&\textbf{ {33}}  \\
\multicolumn{1}{l|}{4D-StOP \cite{kreuzberg2022stop}}&12.50&5&\multicolumn{1}{c|}{31}&17.6&10&31  \\
\multicolumn{1}{l|}{4D-StOP \cite{kreuzberg2022stop} + TTP}&13.63&6&\multicolumn{1}{c|}{31}&18.46&11&31  \\ \hline
\multicolumn{1}{l|}{\textbf{NextStop (Ours)}}&14.5&7&\multicolumn{1}{c|}{30}&17.32&10&30    \\ \hline
\smallskip
TTP - \textit{Tuned Tracking Parameter}
\end{tabular}%
}
\caption{Scores of \textit{\strClass} class on The SemanticKITTI Validation Set}
\label{tab:\strClass-mots-lstq-identity}
\end{table}


\renewcommand{\strClass}{Motorcycle}

\begin{table}[ht!]
\resizebox{\columnwidth}{!}{%
\begin{tabular}{lccccccccc}
\multicolumn{7}{c}{\textbf{\strClass}} \\ \hline\hline
\multicolumn{1}{c}{}          & \multicolumn{3}{c}{LSTQ${_1}$}    & \multicolumn{3}{c}{LSTQ${_{50}}$} \\ \hline
\multicolumn{1}{c|}{Method} &
  LSTQ$\uparrow$ &
  \begin{tabular}[c]{@{}c@{}}S\_$assoc$$\uparrow$\end{tabular} &
  \multicolumn{1}{c|}{S\_$cls$$\uparrow$} &
  LSTQ$\uparrow$ &
  \begin{tabular}[c]{@{}c@{}}S\_$assoc$$\uparrow$\end{tabular} &
  S\_$cls$$\uparrow$ \\ \hline
\multicolumn{1}{l|}{4D-PLS$\>\>\>\>\>$\cite{aygun20214d}}&49.41&37&\multicolumn{1}{c|}{66}&50.73&39&66  \\
\multicolumn{1}{l|}{4D-PLS$\>\>\>\>\>$\cite{aygun20214d} + TTP}&49.41&37&\multicolumn{1}{c|}{66}&50.73&39&66  \\
\multicolumn{1}{l|}{4D-StOP \cite{kreuzberg2022stop}}&58.10&45&\multicolumn{1}{c|}{\textbf{ {75}}}&60&48&75 \\
\multicolumn{1}{l|}{4D-StOP \cite{kreuzberg2022stop} + TTP}&58.73&46&\multicolumn{1}{c|}{\textbf{ {75}}}&60.62&49&75  \\ \hline
\multicolumn{1}{l|}{\textbf{NextStop (Ours)}}&\textbf{ {60.62}}&\textbf{ {49}}&\multicolumn{1}{c|}{\textbf{ {75}}}&\textbf{ {61.84}}&\textbf{ {51}}&\textbf{ {75}}     \\ \hline
\smallskip
TTP - \textit{Tuned Tracking Parameter}
\end{tabular}%
}
\caption{Scores of \textit{\strClass} class on The SemanticKITTI Validation Set}
\label{tab:\strClass-mots-lstq-identity}
\end{table}

\renewcommand{\strClass}{Truck}
\begin{table}[ht!]
\resizebox{\columnwidth}{!}{%
\begin{tabular}{lccccccccc}
\multicolumn{7}{c}{\textbf{\strClass}} \\ \hline\hline
\multicolumn{1}{c}{}           & \multicolumn{3}{c}{LSTQ${_1}$}    & \multicolumn{3}{c}{LSTQ${_{50}}$} \\ \hline
\multicolumn{1}{c|}{Method} &
  LSTQ$\uparrow$ &
  \begin{tabular}[c]{@{}c@{}}S\_$assoc$$\uparrow$\end{tabular} &
  \multicolumn{1}{c|}{S\_$cls$$\uparrow$} &
  LSTQ$\uparrow$ &
  \begin{tabular}[c]{@{}c@{}}S\_$assoc$$\uparrow$\end{tabular} &
  S\_$cls$$\uparrow$ \\ \hline
\multicolumn{1}{l|}{4D-PLS$\>\>\>\>\>$\cite{aygun20214d}} &41.42&44&\multicolumn{1}{c|}{39}&51.11&67&39 \\
\multicolumn{1}{l|}{4D-PLS$\>\>\>\>\>$\cite{aygun20214d} + TTP}&41.9&45&\multicolumn{1}{c|}{39}&51.11&67&39   \\
\multicolumn{1}{l|}{4D-StOP \cite{kreuzberg2022stop}}&72.24&60&\multicolumn{1}{c|}{87}&88.48&90&87  \\
\multicolumn{1}{l|}{4D-StOP \cite{kreuzberg2022stop} + TTP}&72.24&60&\multicolumn{1}{c|}{87}&88.97&91&87  \\ \hline
\multicolumn{1}{l|}{\textbf{NextStop (Ours)}}&\textbf{ {76.05}}&\textbf{ {65}}&\multicolumn{1}{c|}{\textbf{ {89}}}&\textbf{ {93.39}}&\textbf{ {98}}&\textbf{ {89}}    \\ \hline
\smallskip
TTP - \textit{Tuned Tracking Parameter}
\end{tabular}%
}
\caption{Scores of \textit{\strClass} class on The SemanticKITTI Validation Set}
\label{tab:\strClass-mots-lstq-identity}
\end{table}


\renewcommand{\strClass}{Other-vehicle}
\begin{table}[ht!]
\resizebox{\columnwidth}{!}{%
\begin{tabular}{lccccccccc}
\multicolumn{7}{c}{\textbf{\strClass}} \\ \hline\hline
\multicolumn{1}{c}{}          & \multicolumn{3}{c}{LSTQ${_1}$}    & \multicolumn{3}{c}{LSTQ${_{50}}$} \\ \hline
\multicolumn{1}{c|}{Method} &
  LSTQ$\uparrow$ &
  \begin{tabular}[c]{@{}c@{}}S\_$assoc$$\uparrow$\end{tabular} &
  \multicolumn{1}{c|}{S\_$cls$$\uparrow$} &
  LSTQ$\uparrow$ &
  \begin{tabular}[c]{@{}c@{}}S\_$assoc$$\uparrow$\end{tabular} &
  S\_$cls$$\uparrow$ \\ \hline
\multicolumn{1}{l|}{4D-PLS$\>\>\>\>\>$\cite{aygun20214d}}&35&25&\multicolumn{1}{c|}{49}&35.7&26&49 \\
\multicolumn{1}{l|}{4D-PLS$\>\>\>\>\>$\cite{aygun20214d} + TTP}&35&25&\multicolumn{1}{c|}{49}&35.7&26&49    \\
\multicolumn{1}{l|}{4D-StOP \cite{kreuzberg2022stop}}&49.14&35&\multicolumn{1}{c|}{69}&51.2&38&69 \\
\multicolumn{1}{l|}{4D-StOP \cite{kreuzberg2022stop} + TTP}&49.14&35&\multicolumn{1}{c|}{69}&51.2&38&69  \\ \hline
\multicolumn{1}{l|}{\textbf{NextStop (Ours)}}&\textbf{ {60.66}}&\textbf{ {46}}&\multicolumn{1}{c|}{\textbf{ {80}}}&\textbf{ {63.24}}&\textbf{ {50}}&\textbf{ {80}} \\ \hline
\smallskip
TTP - \textit{Tuned Tracking Parameter}
\end{tabular}%
}
\caption{Scores of \textit{\strClass} class on The SemanticKITTI Validation Set}
\label{tab:\strClass-mots-lstq-identity}
\end{table}


\clearpage
\renewcommand{\strClass}{Person}
\begin{table}[ht!]
\resizebox{\columnwidth}{!}{%
\begin{tabular}{lccccccccc}
\multicolumn{7}{c}{\textbf{\strClass}} \\ \hline\hline
\multicolumn{1}{c}{}             & \multicolumn{3}{c}{LSTQ${_1}$}    & \multicolumn{3}{c}{LSTQ${_{50}}$} \\ \hline
\multicolumn{1}{c|}{Method} &
  LSTQ$\uparrow$ &
  \begin{tabular}[c]{@{}c@{}}S\_$assoc$$\uparrow$\end{tabular} &
  \multicolumn{1}{c|}{S\_$cls$$\uparrow$} &
  LSTQ$\uparrow$ &
  \begin{tabular}[c]{@{}c@{}}S\_$assoc$$\uparrow$\end{tabular} &
  S\_$cls$$\uparrow$ \\ \hline
\multicolumn{1}{l|}{4D-PLS$\>\>\>\>\>$\cite{aygun20214d}}&26.98&14&\multicolumn{1}{c|}{52}&33.82&22&52\\
\multicolumn{1}{l|}{4D-PLS$\>\>\>\>\>$\cite{aygun20214d} + TTP}&27.92&15&\multicolumn{1}{c|}{52}&33.82&22&52   \\
\multicolumn{1}{l|}{4D-StOP \cite{kreuzberg2022stop}}&41.83&25&\multicolumn{1}{c|}{70}&54.86&43&70\\
\multicolumn{1}{l|}{4D-StOP \cite{kreuzberg2022stop} + TTP}&44.27&28&\multicolumn{1}{c|}{70}&55.49&44&70\\ \hline
\multicolumn{1}{l|}{\textbf{NextStop (Ours)}}&\textbf{ {52.66}}&\textbf{ {38}}&\multicolumn{1}{c|}{\textbf{ {73}}}&\textbf{ {59.19}}&\textbf{ {48}}&\textbf{ {73}} \\ \hline
\smallskip
TTP - \textit{Tuned Tracking Parameter}
\end{tabular}%
}
\caption{Scores of \textit{\strClass} class on The SemanticKITTI Validation Set}
\label{tab:\strClass-mots-lstq-identity}
\end{table}


\renewcommand{\strClass}{Bicyclist}
\begin{table}[ht!]
\resizebox{\columnwidth}{!}{%
\begin{tabular}{lcccccc}
\multicolumn{7}{c}{\textbf{\strClass}} \\ \hline\hline
\multicolumn{1}{c}{}          & \multicolumn{3}{c}{LSTQ${_1}$}    & \multicolumn{3}{c}{LSTQ${_{50}}$} \\ \hline
\multicolumn{1}{c|}{Method} &
  LSTQ$\uparrow$ &
  \begin{tabular}[c]{@{}c@{}}S\_$assoc$$\uparrow$\end{tabular} &
  \multicolumn{1}{c|}{S\_$cls$$\uparrow$} &
  LSTQ$\uparrow$ &
  \begin{tabular}[c]{@{}c@{}}S\_$assoc$$\uparrow$\end{tabular} &
  S\_$cls$$\uparrow$ \\ \hline
\multicolumn{1}{l|}{4D-PLS$\>\>\>\>\>$\cite{aygun20214d}}&36.94&15&\multicolumn{1}{c|}{\textbf{ {91}}}&69.44&53&\textbf{ {91}}\\
\multicolumn{1}{l|}{4D-PLS$\>\>\>\>\>$\cite{aygun20214d} + TTP}&34.40&13&\multicolumn{1}{c|}{\textbf{ {91}}}&64&45&\textbf{ {91}}  \\
\multicolumn{1}{l|}{4D-StOP \cite{kreuzberg2022stop}}&39.8&18&\multicolumn{1}{c|}{88}&75.04&64&88\\
\multicolumn{1}{l|}{4D-StOP \cite{kreuzberg2022stop} + TTP}&39.8&18&\multicolumn{1}{c|}{88}&74.45&63&88\\ \hline
\multicolumn{1}{l|}{\textbf{NextStop (Ours)}}&\textbf{ {45.5}}&\textbf{ {23}}&\multicolumn{1}{c|}{90}&\textbf{ {76.48}}&\textbf{ {65}}&90 \\ \hline
\smallskip
TTP - \textit{Tuned Tracking Parameter}
\end{tabular}%
}
\caption{Scores of \textit{\strClass} class on The SemanticKITTI Validation Set}
\label{tab:\strClass-mots-lstq-identity}
\end{table}

%% file: figures2/Qualitative_Results/Successes/moving-car2.tex
\begin{figure} [ht!]

\subfloat[4D-STOP Tracking Results]{ \frame{
\includegraphics[width=1\textwidth,height=1\textheight,keepaspectratio]
{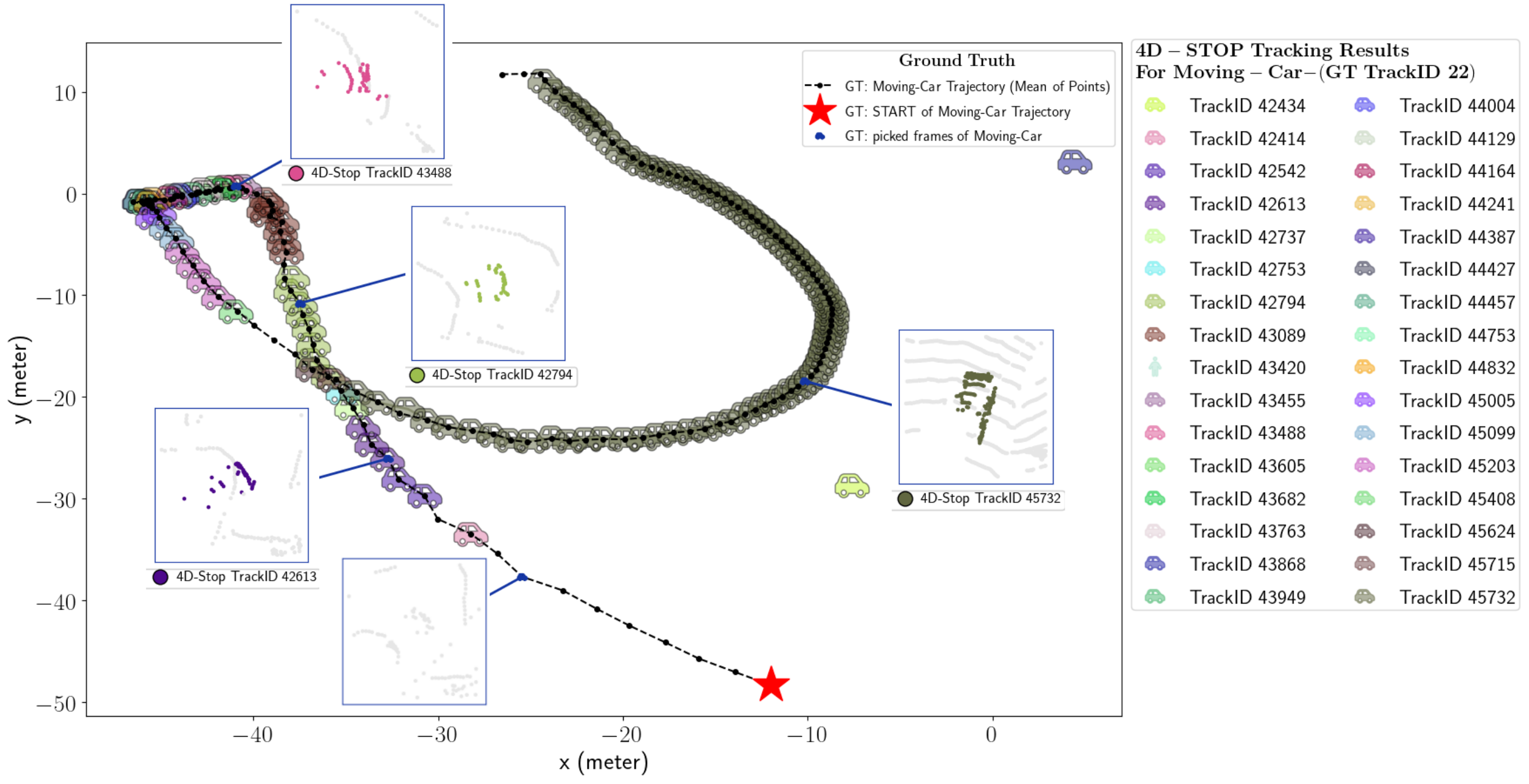}}
\label{moving-car2_stop} }  

\subfloat[NextStop (ours) Tracking Results]{ \frame{
\includegraphics[width=1\textwidth,height=1\textheight,keepaspectratio]
{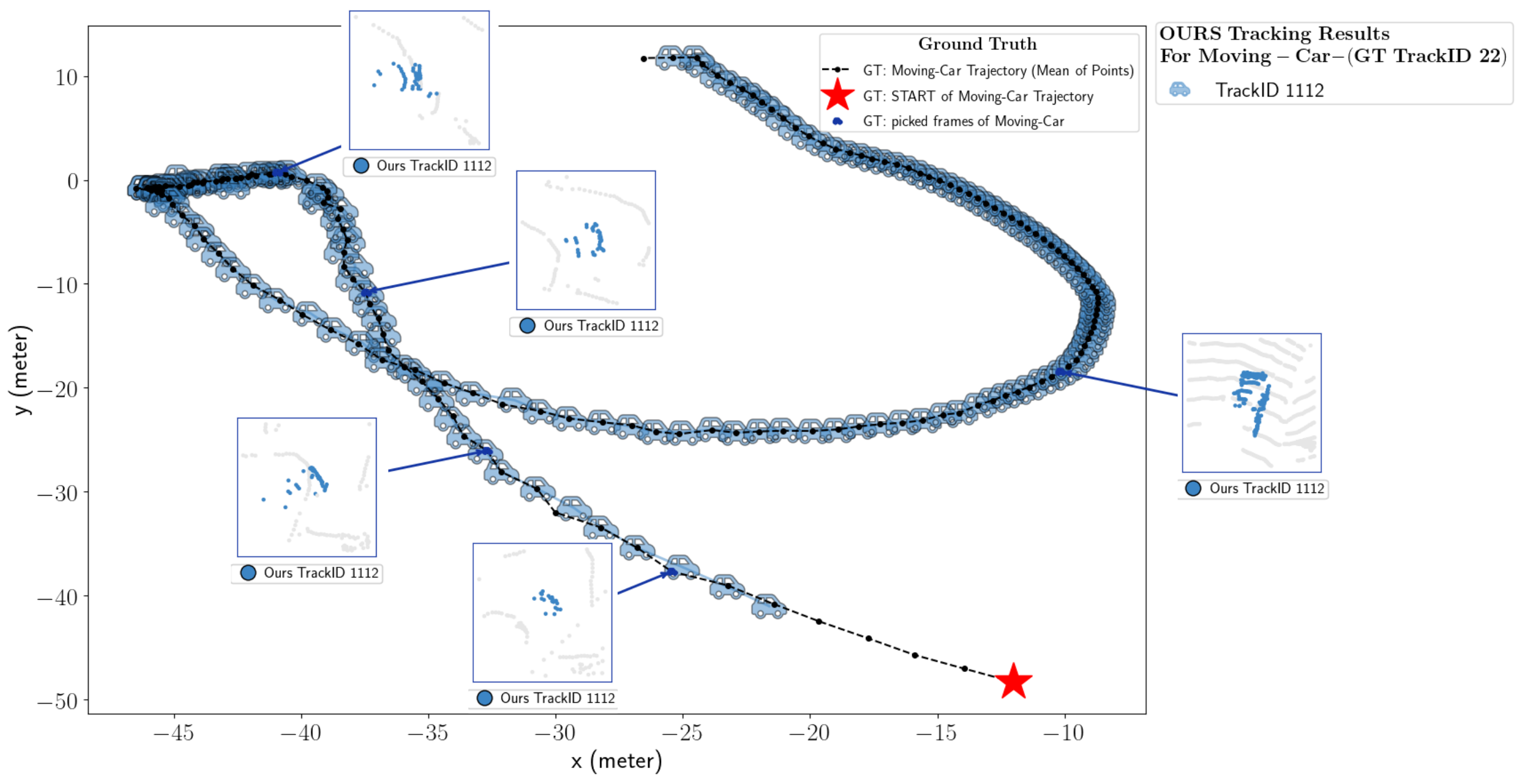}}
\label{moving-car2_ours}}

\small
\begin{flushleft}
{
\textbf{Efficient Tracking of a \textit{Moving Car} with Reduced ID-Switches:}
Presenting the tracked trajectory of a single \textit{Moving Car} object, we accomplished a reduction in ID-Switches from 32 distinct track IDs in the 4D-Stop result (see \autoref{moving-car2_stop}) to a single track ID in our NextStop tracker (see \autoref{moving-car2_ours}). Furthermore, our tracking starts earlier, capturing the object when it is distant and small-sized, resulting in improved trajectory coverage.
}
\end{flushleft}
\medskip

\caption{Visualization of Tracking Results of a \textit{Moving Car}} 
\label{fig:Qualitative_Results_moving-car2_example}
\end{figure}

%% file: figures2/Qualitative_Results/Successes/moving-person.tex
\begin{figure} [ht!]

\subfloat[4D-STOP Tracking Results]{ \frame{
\includegraphics[width=1\textwidth,height=1\textheight,keepaspectratio]
{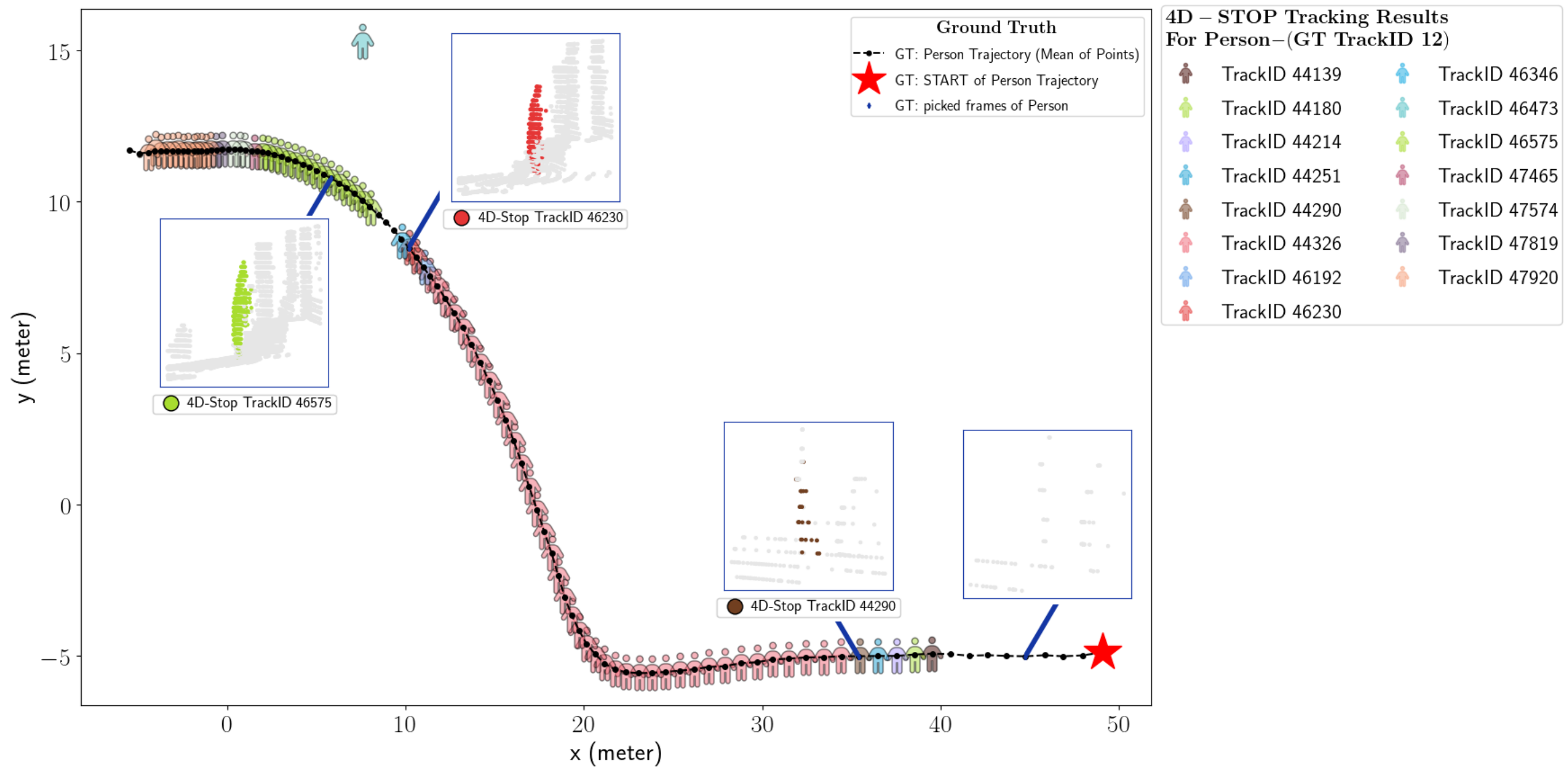}}
\label{moving-person.12_stop} }  

\subfloat[NextStop (ours) Tracking Results]{ \frame{
\includegraphics[width=1\textwidth,height=1\textheight,keepaspectratio]
{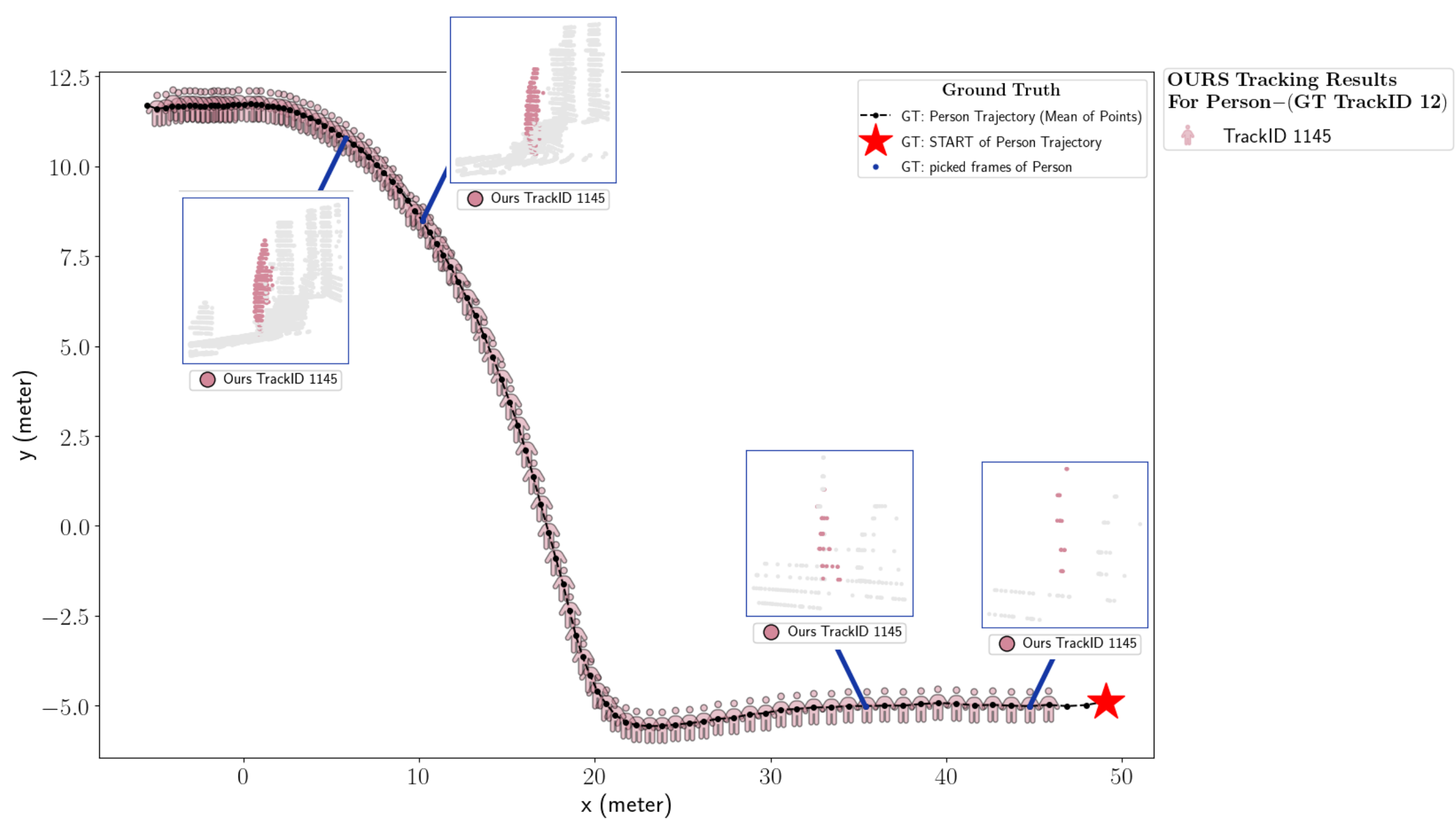}}
\label{moving-person.12_ours}}

\small
\begin{flushleft}
{
\textbf{Illustration of tracking a small-size object:} In this scenario, we are tracking the movements of a \textit{Moving Person}. The person's relative trajectory begins at a considerable distance (approximately 50 meters on the x-axis) and gradually converges towards closer proximity. Initially, when the object is distant, its size measures 22 points, increasing gradually as it approaches, peaking at 290 points. Our approach (\autoref{moving-person.12_ours}) exhibits the earliest detection and maintains a single track ID, keeping  the longest tracking coverage throughout the entire trajectory of this small object.
}

\end{flushleft}
\medskip

\caption{Visualization of Tracking Results of a Small-Sized Object from the \textit{Moving Person} Class} 
\label{fig:Qualitative_Results_moving-person.12_example}
\end{figure}

%% file: figures2/Qualitative_Results/Successes/solution_motivation.tex
\begin{figure}[ht!]
\includegraphics[width=1\textwidth,height=1\textheight,keepaspectratio]
{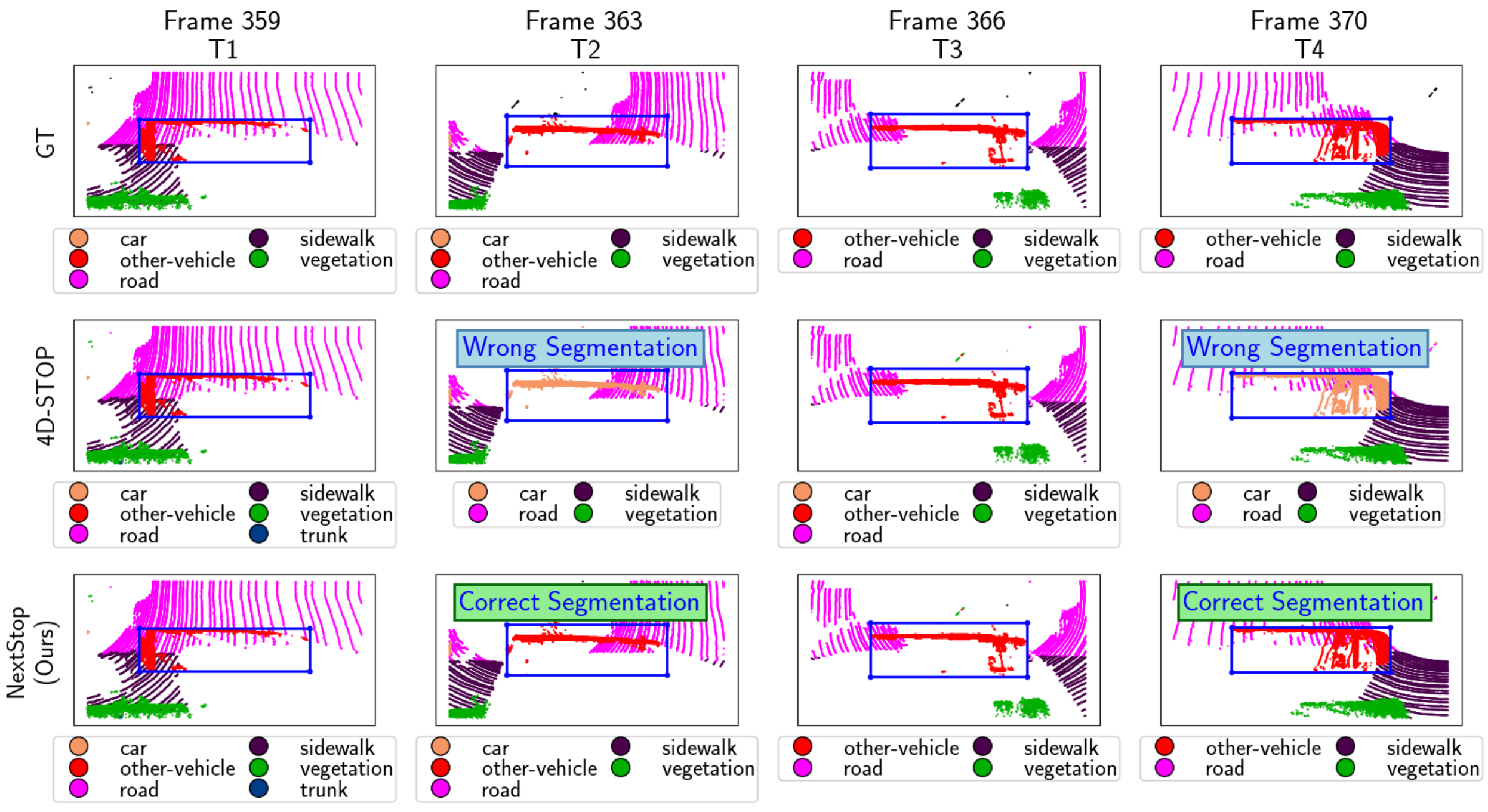}

\small
\begin{flushleft}
{
\textbf{Improving Segmentation Results with Tracking Data:}
Highlighting semantic segmentation, we provide a bird's-eye view of the \textit{other-vehicle} object. The top row shows Ground Truth (GT), the middle row displays segmentation results from 4D-Stop \cite{kreuzberg2022stop}, and the bottom row presents the improved results from NextStop (ours). Each column represents a moment in time progressing from left to right, with a blue bounding box surrounding the object of interest. The incorporation of tracking information accumulated across frames aided in rectifying minor segmentation errors.
}

\end{flushleft}
\medskip

\caption{Visualization of Improved Segmentation Results} 

\label{fig:Qualitative_Results_solution_motivation_example}
\end{figure}

%% file: figures2/Qualitative_Results/Successes/moving-car.tex
\begin{figure}[ht!]
\subfloat[4D-STOP \cite{kreuzberg2022stop} Tracking Results]{ \frame{
\includegraphics[width=1\textwidth,height=1\textheight,keepaspectratio]
{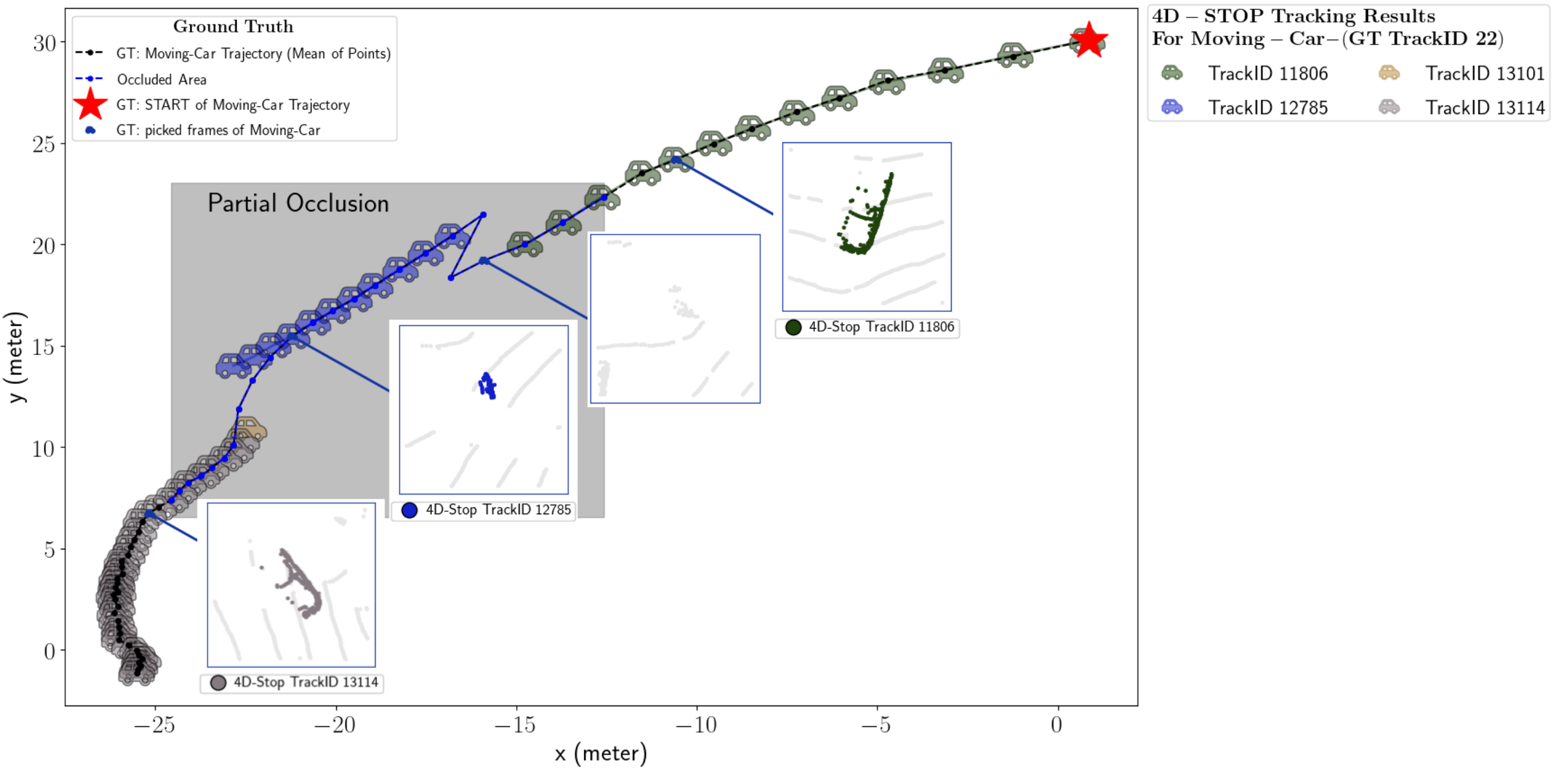}}
\label{gt_class 252_gt_id_22_stop} }  \\


\subfloat[NextStop (ours) Tracking Results]{ \frame{
\includegraphics[width=1\textwidth,height=1\textheight,keepaspectratio]
{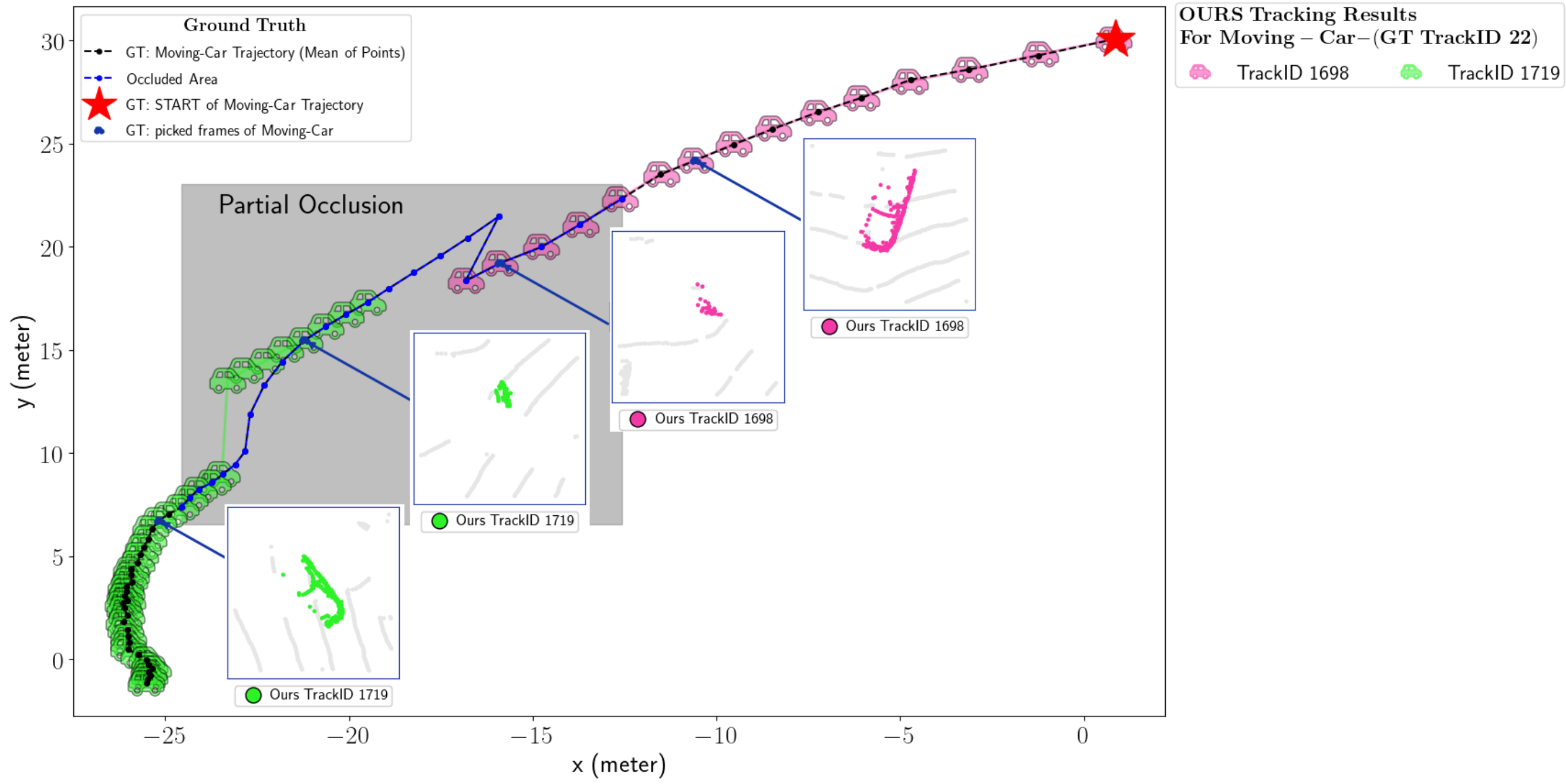}}
\label{gt_class 252_gt_id_22_ours}}

\small
\begin{flushleft}
{
\textbf{Partial Improvement in Tracking \textit{Moving Car} under Occlusion:}
During partial occlusion, the \textit{Moving Car} being tracked undergoes significant orientation changes, posing a challenge for both our approach (\autoref{gt_class 252_gt_id_22_ours}) and the 4D-STOP method \cite{kreuzberg2022stop} (\autoref{gt_class 252_gt_id_22_stop}) to maintain its consistent identity. In this challenging scenario, our NextStop tracker demonstrated partial improvement. While our tracker successfully associates the identity observed during occlusion with the post-occlusion identity, resulting in fewer ID switches compared to 4D-STOP, there are nuances. Notably, during the occlusion, 4D-STOP tracked the object earlier, as NextStop avoids tracking every new detection to avoid false positives of tracklets.
}

\end{flushleft}
\medskip

\caption{Visualizing Partial Improvement in Tracking \textit{Moving Car} under Occlusion} 
\label{fig:Qualitative_Results_gt_class 252_gt_id_22_example}
\end{figure}

%% file: sections/conclusion.tex
\section{Conclusion}
We presented a multi-object tracking algorithm that utilizes data from a 3D point cloud captured by a LiDAR sensor, named NextStop. It uses a \textit{tracking-by-detection} approach, where the detection is taken from an off-the-shelf network. In addition,  our NextStop tracker possesses two primary attributes:
(i) First, we incorporate motion estimators (based on the Kalman filter) to minimize the impact of incorrect detections on the tracker. Since each object has different properties, we divided the estimators into categories of classes: \textit{vehicles}, \textit{bikes}, and \textit{pedestrians}.
(ii) Second, we implemented two priority mechanisms: tracklet prioritization and detection prioritization. Tracklet prioritization prioritizes trustworthy tracklets over untrustworthy ones, whereas detection prioritization prioritizes high-scoring detection results over low-scoring ones. 

We demonstrated that our NextStop tracker offers greater continuity, with fewer ID switches and earlier tracking initiation compared to alternative approaches. The tracker showed substantial improvement in the LSTQ metric, which consists of association and classification scores. Our analysis reveals notable enhancements in the association score, particularly benefiting small-sized classes and objects like "person" and "bicyclist." Additionally, using tracking information to correct temporal inconsistencies in semantic segmentation improved the classification score. However, despite utilizing motion estimators, inaccurate detections, such as those for the "bicycle" class, can cause our tracker to deviate.

Overall, we hold the belief that effective detection plays a crucial role in achieving success in the \textit{track-by-detection} approach. Our tracker is designed to minimize the effects of imprecise detection. It includes integrated components that instruct the tracker on when, how, and to what extent to rely on the detection prediction or its motion estimation results. 

In the future, there is potential for further investigation into the exploration of a more mutual relationship between detection and tracking. While the conventional approach often utilizes detection techniques for tracking, implementing an approach, wherein information obtained from the tracker is utilized to enhance detection and vice versa, has the potential to yield favorable outcomes. This dual feedback mechanism necessitates additional investigation and scholarly inquiry.

%% file: sections/supplementary.tex

\section{Ablation Study}
\label{sec:Ablation_Study}

The primary objective of our ablation study is to assess the effectiveness of each component of the bounding box tracker and to quantify the contribution of each individual component.


\input{tables/supplementary/no_kf}

\PAR{Motion Estimation Contribution:} In \autoref{tab:ablation_study_no_kf}, we present the ${LSTQ}_{1}$ score obtained with and without Kalman filter motion estimation. It appears that employing the Kalman filter yields superior LSTQ results across all classes.


\PAR{DIoU Contribution:} In \autoref{tab:ablation_study_giou}, we present the ${LSTQ}_{1}$ scores obtained using GIoU as a matching score in the data association block, in comparison to the NextStop baseline, which utilizes DIoU. Notably, both GIoU and DIoU values range from -1 to 1, hence no adjustments were required to thresholds, and we adhered to those set in \autoref{tab:Implementations-parameters}. It appears that DIoU contributed to improvements in the classes \textit{Car}, \textit{Bicycle}, \textit{Person}, and \textit{Bicyclist}, while the remaining classes remained unchanged.

\input{tables/supplementary/gIOU}



\PAR{Tracklets State Contribution:} In \autoref{tab:ablation_study_no_candidate}, we present the ${LSTQ}_{1}$ scores obtained when eliminating the concept of tracklet states, specifically removing the \textit{Candidate} state. It's evident that removing the candidate state, thereby retaining only the active state, resulted in a higher association score, as it eliminates the hiding of any tracklet. However, this alteration also led to a decrease in the classification score, emphasizing the significant contribution of tracklet states to the classification score.

\input{tables/supplementary/no_candidate}



\PAR{Split by Detection Score Contribution:} In \autoref{tab:ablation_study_no_detection_split}, we display the ${LSTQ}_{1}$ scores achieved without splitting the detection into high score and low score detections, compare to the NextStop baseline. It appears that, apart from a slight contribution to the \textit{car} class, dividing the detections by their score did not significantly enhance performance. Conversely, for classes such as \textit{Motorcycle}, \textit{Other-vehicle}, and \textit{Bicyclist}, the impact was reversed.

\input{tables/supplementary/no_detection_split}


\section{Discrete Kalman Filter}
\label{sec:kalman}
\newcommand{\tikzmark}[1]{\tikz[baseline,remember picture] \coordinate (#1) {};}

The Kalman filter \cite{bishop2001introduction} is a mathematical algorithm used in control systems and estimation processes, aiming to provide an optimal estimate of a system's state, $\mathbf{x} \in \mathbb{R}^n$, based on noisy measurements $\mathbf{z} \in \mathbb{R}^m$ and the control input $\mathbf{u} \in \mathbb{R}^l$ over time.

\subsection{The Model} 
\label{sec:appendix_kalmman_the_model}
The model assumes that the true state at time step $k$, denoted as $\mathbf{x}_k$, is derived from the state at time step $k-1$. When control information is absent, the discrete Kalman filter model is describe by \autoref{eqn:kalman_state_model}, and the observation (or measurement) $\mathbf{z}_k$ of the true state $\mathbf{x}_k$ adheres to \autoref{eqn:kalman_observation_model}.

\begin{equation}
\label{eqn:kalman_state_model}
\mathbf{x}_{k} =  \mathbf{F}_{k} \mathbf{x}_{k-1} + \mathbf{w}_{k}
\end{equation}

\begin{equation}
\label{eqn:kalman_observation_model}
\mathbf{z}_{k} =  \mathbf{H}_{k} \mathbf{x}_{k}  + \mathbf{v}_{k}
\end{equation}

The vectors of random variables $\mathbf{w}_{k} \in \mathbb{R}^{n}$ and $\mathbf{v}_{k} \in \mathbb{R}^{m}$ denote the process and measurement noise, respectively. They are assumed to be mutually independent of each other. Additionally, each vector is independently and identically distributed (IID), following Gaussian white distributions , i.e., 
$\mathbf{w}_{k} \sim \mathcal{N}(\mathbf{0},\,\mathbf{Q}_{k})$ and $\mathbf{v}_{k} \sim \mathcal{N}(\mathbf{0},\,\mathbf{R}_{k})\,$.

In Equation \ref{eqn:kalman_state_model}, 
$\mathbf{F}_{k} \in \mathbb{R}^{nxn}$  represents the state transition matrix. This matrix establishes the connection between the previous state $\mathbf{x}_{k-1}$ at time step $k-1$, and the current state $\mathbf{x}_{k}$ at time step $k$.

In Equation \ref{eqn:kalman_observation_model}, $\mathbf{H}_{k} \in \mathbb{R}^{mxn}$ represents the observation matrix. This matrix connects the state $\mathbf{x}_{k}$ to the measurement $\mathbf{z}_k$ at time step $k$.

\subsection{The Equations} 
\label{sec:appendix_kalmman_the_Equations}
The Kalman filter is typically divided into two main phases: \textit{Prediction} and \textit{Update} (also known as \textit{Correction}).
At each step $k$, the \textit{Prediction} equations \autoref{eqn:kalman_predict} are used to predict the prior estimate state $\mathbf{\bar{x}}_{k|k-1}$ and the prior covariance matrix $\mathbf{\bar{P}}_{k|k-1}$. Then, if a measurement $\mathbf{z}_{k}$ has been observed, the \textit{Update} equations \autoref{eqn:kalman_update} are used to predict the posteriori state estimate $\mathbf{\hat{x}}_{k|k}$ and the posteriori covariance matrix $\mathbf{\hat{P}}_{k|k}$. If no measurement has been observed, the \textit{Prediction} equations are utilized to project the state forward until the next scheduled measurement is obtained.  

\begin{equation}
\label{eqn:kalman_predict}
\begin{aligned}
\text{Predicted (a priori) state estimate: }  & \mathbf{\bar{x}}_{k|k-1} =  \mathbf{F}_{k} \mathbf{\hat{x}}_{k-1|k-1}
\\
\text{Predicted (a priori) estimate covariance: } & \mathbf{\bar{P}}_{k|k-1} =  \mathbf{F}_{k} \mathbf{\hat{P}}_{k-1|k-1} {{\mathbf{F}^T_{k}} + \mathbf{Q}_{k}}
\end{aligned}
\end{equation}

\begin{equation}
\label{eqn:kalman_update}
\begin{aligned}
\text{Innovation: } & \mathbf{\tilde{y}}_{k} = \mathbf{z}_{k} - \mathbf{H}_{k} \mathbf{\bar{x}}_{k|k-1} 
\\
\text{Innovation covariance: } & \mathbf{S}_{k} = \mathbf{H}_{k} \mathbf{\bar{P}}_{k|k-1}{\mathbf{H}^T_{k}} + {\mathbf{R}_{k}}
\\
\text{Kalman gain: } & \mathbf{K}_{k} = \mathbf{\bar{P}}_{k|k-1} {\mathbf{H}^T_{k}}{\mathbf{S}^{-1}_{k}}
\\
\text{Updated (a posteriori) state estimate: } & \mathbf{\hat{x}}_{k|k} = \mathbf{\hat{x}}_{k|k-1} + \mathbf{K}_{k} \mathbf{\tilde{y}}_{k}
\\
\text{Updated (a posteriori) estimate covariance: } & \mathbf{\hat{P}}_{k|k} = (\mathbf{I}-\mathbf{K}_{k} \mathbf{H}_{k}) \mathbf{\bar{P}}_{k|k-1} 
\end{aligned}
\end{equation}

\subsection{Implementation Details} 
\label{sec:appendix_kalmman_Implementation_details}

As the SemmanticKITTI dataset \cite{behley2019iccv} lacks control information, the model equation \autoref{eqn:kalman_state_model} becomes:

\begin{equation}
\label{eqn:kalman_state_model_Implementation}
\mathbf{x}_{k} =  \mathbf{F}_{k} \mathbf{x}_{k-1} + \mathbf{w}_{k}
\end{equation}

while the \autoref{eqn:kalman_observation_model} remains the same. We assumed that the objects move with a constant velocity over time, and as such our state vector is represented by \autoref{eqn:kalman_implementation_state}:

\begin{equation}
\label{eqn:kalman_implementation_state}
\mathbf{x}_{k}=\left[ c_x, c_y, c_z, \theta, l, w, h, v_x, v_y, v_z \right]^T
\end{equation}

Where the first 7 elements ${(c_x, c_y, c_z, \theta, l, w, h)}$
represent the 3D bounding box: centroid, orientation, and dimension measurements, and the last elements ${(v_x, v_y, v_z)}$ represents the 3D bounding box velocity.

The measurement vector $\mathbf{z}_{k}$ is represented by \autoref{eqn:kalman_implementation_measurement}:

\begin{equation}
\label{eqn:kalman_implementation_measurement}
\mathbf{z}_{k}=\left[ c_x, c_y, c_z, \theta, l,w, h \right]^T
\end{equation}

Which includes the centroid, orientation, and dimensions of the 3D detection bounding box. 
\bigskip
We picked all matrices $\mathbf{F}_{k}$, $\mathbf{H}_{k}$, $\mathbf{Q}_{k}$ and $\mathbf{R}_{k}$ to be time independent meaning:

\begin{equation*}
\label{eqn:kalman_Implementation_matrices}
\begin{aligned}
\text{State transition matrix: } & \mathbf{F}_{k} =  \mathbf{F}\quad \forall k
\\
\text{Observation matrix : } & \mathbf{H}_{k} =  \mathbf{H}\quad \forall k
\\
\text{Process noise covariance: } & \mathbf{Q}_{k} =  \mathbf{Q}\quad \forall k
\\
\text{Noise measurement covariance: } & \mathbf{R}_{k} =  \mathbf{R}\quad \forall k
\end{aligned}
\end{equation*}

The state transition matrix $\mathbf{F}$ is shown in \autoref{eqn:state transition matrix}, while the observation matrix $\mathbf{H}$ is displayed in  \autoref{eqn:observation matrix}. The values for the process noise covariance $\mathbf{Q}$, the noise measurement covariance $\mathbf{R}$, and the initial state covariance $\mathbf{P}_{0|0}$ were determined and are presented in \autoref{tab:Kalman Filter Implementations-parameters}. To compensate for the SemanticKITTI database's lack of orientation data, we specified Kalman matrices $\mathbf{Q}$ and $\mathbf{R}$ to reduce the orientation component's influence on state and measurement vectors.

For the initial condition of the state vector $\mathbf{x}_{0|0}$ we chose the first 3D detection box that was associate with this tracker, with velocity of zero. In addition, we corrected some of the measurement vector elements $\mathbf{z}_{k}$ that did not have zero mean by adding an offset, as indicated in \autoref{tab:Kalman Filter Implementations-parameters}.

\begin{table}[ht!]
{
\resizebox{0.8\textwidth}{!}{%
{
\begin{tabular}{cllll}
 & \multicolumn{4}{c}{Kalman Parameters} \\\\
 & \multicolumn{1}{c}{$\mathbf{P}_{0|0}$} & \multicolumn{1}{c}{$\mathbf{Q}$} & \multicolumn{1}{c}{$\mathbf{R}$} & \multicolumn{1}{c}{Z Offset}\\ \hline
Vehicles & \begin{tabular}[c]{@{}l@{}}$\!\begin{aligned}[t]
diag(&10, 10, 10,\\&10, 10, 10,\\&10, 10^4, 10^4,\\&10^4)
\end{aligned}$\end{tabular} & \begin{tabular}[c]{@{}l@{}}$\!\begin{aligned}[t]
diag(&0,  0,  0,\\& 1, 1, 1,\\& 0.3, 0.01, 0.01,\\& 0.01)
\end{aligned}$\end{tabular} & \begin{tabular}[c]{@{}l@{}}$\!\begin{aligned}[t]
diag(&0.1, 0.1,\\& 0.1, 10^4,\\& 0.1, 0.1,\\& 0.1)
\end{aligned}$\end{tabular} & \begin{tabular}[c]{@{}l@{}} $c_z$ += 0.05\\ $h$ -= 0.1\\ \end{tabular}\\ \hline

Bikes & \begin{tabular}[c]{@{}l@{}}$\!\begin{aligned}[t]
diag(&10, 10, 10,\\& 10, 10, 10,\\& 10, 10^4, 10^4,\\& 10^4)
\end{aligned}$\end{tabular} & \begin{tabular}[c]{@{}l@{}}$\!\begin{aligned}[t]
diag(&0,  0,  0,\\& 1, 1, 1,\\& 0.3, 0.01, 0.01,\\& 0.01)
\end{aligned}$
\end{tabular} & \begin{tabular}[c]{@{}l@{}}$\!\begin{aligned}[t]
diag(&0.1, 0.1,\\& 0.1, 10^4,\\& 0.1, 0.1,\\& 0.1)
\end{aligned}$\end{tabular} & \begin{tabular}[c]{@{}l@{}} $c_z$ -= 0.025\\ $h$ += 0.0625\\ \end{tabular}\\ \hline

Pedestrian & \begin{tabular}[c]{@{}l@{}}$\!\begin{aligned}[t]
diag(&10, 10, 10,\\& 10, 10, 10,\\& 10, 10^4, 10^4,\\& 10^4)
\end{aligned}$
\end{tabular} & \begin{tabular}[c]{@{}l@{}}$\!\begin{aligned}[t]
diag(&0,  0,  0,\\& 1, 0.4, 0.4,\\& 0.4, 0.01, 0.01,\\& 0.01)
\end{aligned}$
\end{tabular} & \begin{tabular}[c]{@{}l@{}}$\!\begin{aligned}[t]
diag(&0.1, 0.1,\\& 0.1, 10^4,\\& 0.1, 0.1,\\& 0.1)
\end{aligned}$\end{tabular}
& \begin{tabular}[c]{@{}l@{}} $c_z$ += 0.028125\\ $h$ -= 0.1\\ \end{tabular}
\end{tabular}%
}
}
}
\caption{Kalman Filter Implementations  Parameters}
\label{tab:Kalman Filter Implementations-parameters}
\end{table}



\begin{equation}
F = 
\begin{bmatrix}
1&0&0&0&0&0&0&1&0&0\\
0&1&0&0&0&0&0&0&1&0\\
0&0&1&0&0&0&0&0&0&1\\
0&0&0&1&0&0&0&0&0&0\\  
0&0&0&0&1&0&0&0&0&0\\
0&0&0&0&0&1&0&0&0&0\\
0&0&0&0&0&0&1&0&0&0\\
0&0&0&0&0&0&0&1&0&0\\
0&0&0&0&0&0&0&0&1&0\\
0&0&0&0&0&0&0&0&0&1\\
\end{bmatrix}
_{10x10}
\label{eqn:state transition matrix}
\end{equation}

\begin{equation}
H = 
\begin{bmatrix}
1&0&0&0&0&0&0&0&0&0\\      
0&1&0&0&0&0&0&0&0&0\\
0&0&1&0&0&0&0&0&0&0\\
0&0&0&1&0&0&0&0&0&0\\
0&0&0&0&1&0&0&0&0&0\\
0&0&0&0&0&1&0&0&0&0\\
0&0&0&0&0&0&1&0&0&0\\
\end{bmatrix}
_{7x7}
\label{eqn:observation matrix}
\end{equation}

%% file: tables/supplementary/no_kf.tex
\begin{table}[h!]
\resizebox{\columnwidth}{!}{%
\begin{tabular}{ccccccc}
    & \multicolumn{3}{c}{\begin{tabular}[c]{@{}c@{}}NextStop\\ Without Motion Estimation\end{tabular}} & \multicolumn{3}{c}{\begin{tabular}[c]{@{}c@{}}Baseline: NextStop (Ours)\\ With Motion Estimation\end{tabular}} \\ \hline
\multicolumn{1}{c|}{class} & $LSTQ_{1}\uparrow$ & S\_$assoc$$\uparrow$ & \multicolumn{1}{c|}{S\_$cls$$\uparrow$} & $LSTQ_{1}\uparrow$ & S\_$assoc$$\uparrow$& S\_$cls$$\uparrow$\\ \hline
\multicolumn{1}{c|}{Things} & 63.54 & 60.73 & 66.5 & \multicolumn{1}{|c}{\textbf{65.28}}  & \textbf{63.84} & \textbf{66.76} \\ \hline
\multicolumn{1}{c|}{Car} & 87.53 & 79 & 97 & \multicolumn{1}{|c}{\textbf{89.64}} & \textbf{82} & \textbf{98} \\
\multicolumn{1}{c|}{Bicycle} & 14.24 & 7 &  29& \multicolumn{1}{|c}{\textbf{14.5}} & \textbf{7} & \textbf{30} \\
\multicolumn{1}{c|}{Motorcycle} & 60.4 & 48 & \textbf{76} & \multicolumn{1}{|c}{\textbf{60.62}} & \textbf{49} & 75 \\
\multicolumn{1}{c|}{Truck} & 75.63 & 65 & 88 & \multicolumn{1}{|c}{\textbf{76.05}} & 65 & \textbf{89} \\
\multicolumn{1}{c|}{Other-vehicle} & 58.95 & 44 & 79 & \multicolumn{1}{|c}{\textbf{60.66}} & \textbf{46} & \textbf{80} \\
\multicolumn{1}{c|}{Person} &  49.48& 31 & \textbf{79} & \multicolumn{1}{|c}{\textbf{52.66}} & \textbf{38} & 73 \\
\multicolumn{1}{c|}{Bicyclist} & 38.61 & 21 & 71&\multicolumn{1}{|c}{\textbf{45.5}} & \textbf{23} & \textbf{90}
\end{tabular}%
}
\caption{Ablation study: Motion Estimation Contribution}

\label{tab:ablation_study_no_kf}

\end{table}


%% file: tables/supplementary/giou.tex
\begin{table}[h!]
\resizebox{\columnwidth}{!}{%
\begin{tabular}{ccccccc}
 & \multicolumn{3}{c}{\begin{tabular}[c]{@{}c@{}}NextStop\\ with GIoU\end{tabular}} & \multicolumn{3}{c}{\begin{tabular}[c]{@{}c@{}}Baseline: NextStop (Ours)\\ with DIoU\end{tabular}} \\ \hline
\multicolumn{1}{c|}{class} & $LSTQ_{1}\uparrow$ & S\_$assoc$$\uparrow$ & \multicolumn{1}{c|}{S\_$cls$$\uparrow$} & $LSTQ_{1}\uparrow$ & S\_$assoc$$\uparrow$& S\_$cls$$\uparrow$\\ \hline
\multicolumn{1}{c|}{Things} & 64.17&61.81&66.62 & \multicolumn{1}{|c}{\textbf{65.28}}  & \textbf{63.84} & \textbf{66.76} \\ \hline
\multicolumn{1}{c|}{Car} & 88.54&80&98 & \multicolumn{1}{|c}{\textbf{89.64}} & \textbf{82} & 98 \\
\multicolumn{1}{c|}{Bicycle} & 13.2&6&29& \multicolumn{1}{|c}{\textbf{14.5}} & \textbf{7} & \textbf{30} \\
\multicolumn{1}{c|}{Motorcycle} & 60.62&49&75 & \multicolumn{1}{|c}{60.62} & 49 & 75 \\
\multicolumn{1}{c|}{Truck} & 76.05&65&89 & \multicolumn{1}{|c}{76.05} & 65 & 89 \\
\multicolumn{1}{c|}{Other-vehicle} & \textbf{61.04}&46&\textbf{81} & \multicolumn{1}{|c}{60.66} & 46 & 80 \\
\multicolumn{1}{c|}{Person} & 49.47&34&72  & \multicolumn{1}{|c}{\textbf{52.66}} & \textbf{38} & \textbf{73} \\
\multicolumn{1}{c|}{Bicyclist} &43.47&21&90 &\multicolumn{1}{|c}{\textbf{45.5}} & \textbf{23} & 90
\end{tabular}%
}
\caption{Ablation study: DIoU Contribution}

\label{tab:ablation_study_giou}

\end{table}


%% file: tables/supplementary/no_candidate.tex
\begin{table}[h!]
\resizebox{\columnwidth}{!}{%
\begin{tabular}{ccccccc}
  & \multicolumn{3}{c}{\begin{tabular}[c]{@{}c@{}}NextStop\\ Without Tracklet State Concept\end{tabular}} & \multicolumn{3}{c}{\begin{tabular}[c]{@{}c@{}}Baseline: NextStop (Ours)\\ With Tracklet State Concept\end{tabular}} \\ \hline
\multicolumn{1}{c|}{class} & $LSTQ_{1}\uparrow$ & S\_$assoc$$\uparrow$ & \multicolumn{1}{c|}{S\_$cls$$\uparrow$} & $LSTQ_{1}\uparrow$ & S\_$assoc$$\uparrow$& S\_$cls$$\uparrow$\\ \hline
\multicolumn{1}{c|}{Things} & 64.04&\textbf{64.46}&63.62  & \multicolumn{1}{|c}{\textbf{65.28}}  & {63.84} & {\textbf{66.76}} \\ \hline
\multicolumn{1}{c|}{Car} &88.72&82&96 & \multicolumn{1}{|c}{\textbf{89.64}} & 82 & \textbf{98} \\
\multicolumn{1}{c|}{Bicycle} &\textbf{16.97}&\textbf{9}&\textbf{32}& \multicolumn{1}{|c}{{14.5}} & {7} & {30} \\
\multicolumn{1}{c|}{Motorcycle} & \textbf{62.62}&\textbf{53}&74 & \multicolumn{1}{|c}{60.62} & 49 & \textbf{75} \\
\multicolumn{1}{c|}{Truck} & 75.04&64&88& \multicolumn{1}{|c}{\textbf{76.05}} & \textbf{65} & \textbf{89} \\
\multicolumn{1}{c|}{Other-vehicle} & 53.54&\textbf{47}&61 & \multicolumn{1}{|c}{\textbf{60.66}} & 46 & \textbf{80} \\
\multicolumn{1}{c|}{Person} & \textbf{53.18}&\textbf{41}&69  & \multicolumn{1}{|c}{52.66} & 38 & \textbf{73} \\
\multicolumn{1}{c|}{Bicyclist} & \textbf{45.95}&\textbf{24}&88&\multicolumn{1}{|c}{45.5} & 23 & \textbf{90}
\end{tabular}%
}
\caption{Ablation study: Tracklets State Contribution}

\label{tab:ablation_study_no_candidate}

\end{table}


%% file: tables/supplementary/no_detection_split.tex
\clearpage
\begin{table}[h!]
\resizebox{\columnwidth}{!}{%
\begin{tabular}{ccccccc}
   & \multicolumn{3}{c}{\begin{tabular}[c]{@{}c@{}}NextStop\\ Without Detection Split by Score\end{tabular}} & \multicolumn{3}{c}{\begin{tabular}[c]{@{}c@{}}Baseline: NextStop (Ours)\\ With Detection Split by Score\end{tabular}} \\ \hline
\multicolumn{1}{c|}{class} & $LSTQ_{1}\uparrow$ & S\_$assoc$$\uparrow$ & \multicolumn{1}{c|}{S\_$cls$$\uparrow$} & $LSTQ_{1}\uparrow$ & S\_$assoc$$\uparrow$& S\_$cls$$\uparrow$\\ \hline
\multicolumn{1}{c|}{Things} &65.21&63.53&\textbf{66.95}  & \multicolumn{1}{|c}{\textbf{65.28}}  & \textbf{63.84} & {66.76} \\ \hline
\multicolumn{1}{c|}{Car} & 89.1&81&98& \multicolumn{1}{|c}{\textbf{89.64}} & \textbf{82} & 98 \\
\multicolumn{1}{c|}{Bicycle} & 14.5&7&30& \multicolumn{1}{|c}{{14.5}} & {7} & {30} \\
\multicolumn{1}{c|}{Motorcycle} & \textbf{61.02}&49&\textbf{76} & \multicolumn{1}{|c}{60.62} & 49 & 75 \\
\multicolumn{1}{c|}{Truck} & 76.05&65&89 & \multicolumn{1}{|c}{76.05} & 65 & 89 \\
\multicolumn{1}{c|}{Other-vehicle} & \textbf{61.7}&\textbf{47}&\textbf{81} & \multicolumn{1}{|c}{60.66} & 46 & 80 \\
\multicolumn{1}{c|}{Person} & 52.66&38&73  & \multicolumn{1}{|c}{52.66} & 38 & 73 \\
\multicolumn{1}{c|}{Bicyclist} & \textbf{46.47}&\textbf{24}&90&\multicolumn{1}{|c}{45.5} & 23 & 90
\end{tabular}%
}
\caption{Ablation study: Split by Detection Score Contribution}

\label{tab:ablation_study_no_detection_split}

\end{table}
